# On the Compilability and Expressive Power of Propositional Planning Formalisms

**Bernhard Nebel** NEBEL@INFORMATIK.UNI-FREIBURG.DE
*Institut für Informatik, Albert-Ludwigs-Universität, Georges-Köhler-Allee, D-79110 Freiburg, Germany*

## Abstract

The recent approaches of extending the GRAPHPLAN algorithm to handle more expressive planning formalisms raise the question of what the formal meaning of "expressive power" is. We formalize the intuition that expressive power is a measure of how concisely planning domains and plans can be expressed in a particular formalism by introducing the notion of "compilation schemes" between planning formalisms. Using this notion, we analyze the expressiveness of a large family of propositional planning formalisms, ranging from basic STRIPS to a formalism with conditional effects, partial state specifications, and propositional formulae in the preconditions. One of the results is that conditional effects cannot be compiled away if plan size should grow only linearly but can be compiled away if we allow for polynomial growth of the resulting plans. This result confirms that the recently proposed extensions to the GRAPHPLAN algorithm concerning conditional effects are optimal with respect to the "compilability" framework. Another result is that general propositional formulae cannot be compiled into conditional effects if the plan size should be preserved linearly. This implies that allowing general propositional formulae in preconditions and effect conditions adds another level of difficulty in generating a plan.

## 1. Introduction

GRAPHPLAN (Blum & Furst, 1997) and SATPLAN (Kautz & Selman, 1996) are among the most efficient planning systems nowadays. However, it is generally felt that the planning formalism supported by these systems, namely, propositional basic STRIPS (Fikes & Nilsson, 1971), is not *expressive* enough. For this reason, much research effort (Anderson, Smith, & Weld, 1998; Gazen & Knoblock, 1997; Kambhampati, Parker, & Lambrecht, 1997; Koehler, Nebel, Hoffmann, & Dimopoulos, 1997) has been devoted in extending GRAPHPLAN in order to handle more powerful planning formalisms such as ADL (Pednault, 1989).

There appears to be a consensus on how much *expressive power* is added by a particular language feature. For example, everybody seems to agree that adding negative preconditions does not add very much to the expressive power of basic STRIPS, whereas conditional effects are considered as a significant increase in expressive power (Anderson et al., 1998; Gazen & Knoblock, 1997; Kambhampati et al., 1997; Koehler et al., 1997). However, it is unclear how to measure the expressive power in a more formal way. Related to this problem is the question of whether "compilation" approaches to extend the expressiveness of a planning formalism are optimal. For example, Gazen and Knoblock (1997) propose a particular method of compiling operators with conditional effects into basic STRIPS operators. This method, however, results in exponentially larger operator sets. While most people (Anderson et al., 1998; Kambhampati et al., 1997; Koehler et al., 1997) agree that we cannot do better than that, nobody has proven yet that a more space-efficient method is impossible.

In order to address the problem of measuring the relative expressive power of planning formalisms, we start with the intuition that a formalism $\mathcal{X}$ is *at least as expressive* as another formalism $\mathcal{Y}$ if *planning domains* and the corresponding *plans* in formalism $\mathcal{Y}$ can be *concisely expressed* in the formalism $\mathcal{X}$. This, at least, seems to be the underlying intuition when expressive power is discussed in the planning literature.

Bäckström (1995) proposed to measure the expressiveness of planning formalisms using his *ESP-reductions*. These reductions are, roughly speaking, polynomial many-one reductions[1] on planning instances that *do not change the plan length*. Using this notion, he showed that all of the propositional variants of basic STRIPS not containing conditional effects or arbitrary logical formulae can be considered as expressively equivalent. However, taking our point of view, ESP-reductions are too restrictive for two reasons. Firstly, plans must have identical size, while we might want to allow a moderate growth. Secondly, requiring that the transformation can be computed in polynomial time is overly restrictive. If we ask for how concisely something can be *expressed*, this does not necessarily imply that there exists a polynomial-time transformation. In fact, one formalism might be as expressive as another one, but the mapping between the formalisms might not be computable at all. This, at least, seems to be the usual assumption made when the term *expressive power* is discussed (Baader, 1990; Cadoli, Donini, Liberatore, & Schaerf, 1996; Erol, Hendler, & Nau, 1996; Gogic, Kautz, Papadimitriou, & Selman, 1995).

Inspired by recent approaches to measure the expressiveness of knowledge representation formalisms (Cadoli et al., 1996; Gogic et al., 1995), we propose to address the questions of how expressive a planning formalism is by using the notion of *compiling* one planning formalism into another one. A compilation scheme from one planning formalism to another differs from a polynomial many-one reduction in that it is not required that the compilation is carried out in polynomial time. However, the result should be expressible in polynomial space. Furthermore, it is required that the operators of the planning instance can be translated *without* considering the initial state and the goal. While this restriction might sound unnecessarily restrictive, it turns out that existing practical approaches to compilation (Gazen & Knoblock, 1997) as well as theoretical approaches (Bäckström, 1995) consider only *structured* transformations where the operators can be transformed independently from the initial state and the goal description. From a technical point of view this restriction guarantees that compilations are non-trivial. If the entire instance could be transformed, a compilation scheme could decide the existence of a plan for the source instance and then generate a small solution-preserving instance in the target formalism, which would lead to the unintuitive conclusion that all planning formalisms have the same expressive power.

As mentioned in the beginning, not only the space taken up by the domain structure is important, but also the space used by the plans. For this reason, we distinguish between compilation schemes in whether they preserve plan size *exactly*, *linearly*, or *polynomially*.

Using the notion of *compilability*, we analyze a wide range of propositional planning formalisms, ranging from basic STRIPS to a planning formalism containing *conditional effects*, *arbitrary boolean formulae*, and *partial state specifications*. As one of the results, we identify two equivalence classes of planning formalisms with respect to *polynomial-time* compilability preserving plan size exactly. This means that adding a language feature to a formalism without leaving the class does not increase the expressive power and should not affect the principal efficiency of

1. We assume that the reader has a basic knowledge of complexity theory (Garey & Johnson, 1979; Papadimitriou, 1994), and is familiar with the notion of *polynomial many-one reductions* and the *complexity classes* P, NP, coNP, and PSPACE. All other notions will be introduced in the paper when needed.

the planning method. However, we also provide results that *separate* planning formalisms using results from computational complexity theory on circuit complexity and non-uniform complexity classes. Such separation results indicate that adding a particular language feature adds to the expressive power and to the difficulty of integrating the feature into an existing planning algorithm. For example, we prove that conditional effects cannot be compiled away and that boolean formulae cannot be compiled into conditional effects—provided the plans in the target formalism are allowed to grow only linearly.

This answers the question posed in the beginning. The compilation approach proposed by Gazen and Knoblock (1997) cannot be more space efficient, even if we allow for linear growth of the plans in the target formalism.[2] Allowing for polynomial growth of the plans, however, the compilation scheme can be more space efficient. Interestingly, it seems to be the case that a compilation scheme that allows for polynomially larger plans is similar to the implementation of conditional effects in the IPP system (Koehler et al., 1997), Kambhampati and colleagues' (1997) planning system, and Anderson and colleagues' (1998) planning system.

The rest of the paper is structured as follows. In Section 2, we introduce the range of propositional planning formalisms analyzed in this paper together with general terminology and definitions. Based on that, we introduce the notion of compilability between planning formalisms in Section 3. In Section 4 we present polynomial-time compilation schemes between different formalisms that preserve the plan size exactly, demonstrating that these formalisms are of identical expressiveness. For all of the remaining cases, we prove in Section 5 that there cannot be any compilation scheme preserving plan size linearly, even if there are no bounds on the computational resources of the compilation process. In Section 6 we reconsider the question of identical expressiveness by using compilation schemes that allow for polynomial growth of the plans. Finally, in Section 7 we summarize and discuss the results.

## 2. Propositional Planning Formalisms

First, we will define a very general propositional planning formalism, which appears almost as expressive as the propositional variant of ADL (Pednault, 1989). This formalism allows for arbitrary boolean formulae as preconditions, conditional effects and partial state specifications. Subsequently, we will specialize this formalism by imposing different syntactic restrictions.

### 2.1 A General Propositional Planning Formalism

Let $\mathbf{\Sigma}$ be the countably infinite set of **propositional atoms** or **propositional variables**. Finite subsets of $\mathbf{\Sigma}$ are denoted by $\Sigma$. Further, $\widehat{\Sigma}$ is defined to be the set consisting of the constants $\top$ (denoting truth) and $\bot$ (denoting falsity) as well as atoms and negated atoms, i.e., the **literals**, over $\Sigma$. The **language of propositional logic** over the logical connectives $\wedge, \vee$, and $\neg$ and the propositional atoms $\Sigma$ is denoted by $\mathcal{L}_\Sigma$. A **clause** is a disjunction of literals. Further, we say that a formula $\varphi \in \mathcal{L}_\Sigma$ is in **conjunctive normal form** (CNF) if it is a conjunction of clauses. It is in **disjunctive normal form** (DNF) if it is a disjunction of conjunctions of literals.

Given a set of literals $L$, by $pos(L)$ we refer to the **positive literals** in $L$, by $neg(L)$ we refer to the **negative literals** in $L$, and by $\sigma(L)$ to the atoms used in $L$, i.e., $\sigma(L) = \{p \in \Sigma \,|\, p \in L$ or $\neg p \in$

2. Note that Gazen and Knoblock's (1997) translation scheme also generates planning operators that depend on the initial state and the goal description. However, these operators simply code the initial state and the goal description and do nothing else. For this reason, we can ignore them here.

$L\}$. Further, we define $\neg L$ to be the **element-wise negation** of $L$, i.e.,

$$\neg L = \{p \mid \neg p \in L\} \cup \{\neg p \mid p \in L\}.$$

A **state** $s$ is a *truth-assignment* for the atoms in $\Sigma$. In the following, we also identify a state with the set of atoms that are true in this state. A **state specification** $S$ is a subset of $\widehat{\Sigma}$, i.e., it is a *logical theory* consisting of literals only. It is called **consistent** iff it does not contain complementary literals or $\perp$. In general, a state specification describes many states, namely all those that satisfy $S$, which are denoted by $Mod(S)$. Only in case that $S$ is **complete**, i.e., for each $p \in \Sigma$ we have either $p \in S$ or $\neg p \in S$, $S$ has precisely one model, namely $pos(S)$. By abusing notation, we will refer to the **inconsistent** state specification by $\perp$, which is the **"illegal" state specification**.

**Operators** are pairs $o = \langle pre, post\rangle$. We use the notation $pre(o)$ and $post(o)$ to refer to the first and second part of an operator $o$, respectively. The **precondition** *pre* is an element of $\mathbf{2}^{\mathcal{L}_\Sigma}$, i.e., it is a set of propositional formulae. The set *post*, which is the set of **postconditions**, consists of **conditional effects**, each having the form

$$\Gamma \Rightarrow L,$$

where the elements of $\Gamma \subseteq \mathcal{L}_\Sigma$ are called **effect conditions** and the elements of $L \subseteq \widehat{\Sigma}$ are called **effects**. If $\Gamma$ or $L$ are singleton sets, e.g., $\{p\} \Rightarrow \{q\}$, we often omit the curly brackets and write $p \Rightarrow q$.

**Example 1** *In order to illustrate the various notions, we will use as a running example planning problems connected with the production of camera-ready manuscripts from LaTeX source files—somewhat simplified, of course. As the set of atoms $\Sigma$, we choose the following set:*

$$\Sigma = \{\mathsf{tex}, \mathsf{aux}, \mathsf{dvi}, \mathsf{log}, \mathsf{ps}, \mathsf{bib}, \mathsf{bbl}, \mathsf{blg}, \mathsf{ind}, \mathsf{idx}, \mathsf{ilg} \\ \mathsf{dvi_ind_ok}, \mathsf{dvi_cite_ok}\}.$$

*These propositional atoms have the following intended meaning. The atoms in the first line represent the presence of the corresponding files, and the atoms in the second line signify that the index and citations are correct in the dvi-file. Based on that, we define the following operators:* bibtex, latex, makeindex. *The first of these operators is very simple. The precondition for its execution is that a* bib- *and an* aux-*file exist. After the successful execution, a* bbl- *and a* blg-*file will have been produced:*

$$\mathsf{bibtex} = \Big\langle \Big\{\mathsf{aux}, \mathsf{bib}\Big\}, \Big\{\top \Rightarrow \{\mathsf{bbl}, \mathsf{blg}\}\Big\}\Big\rangle.$$

*The* makeindex *operator is similar:*

$$\mathsf{makeindex} = \Big\langle \Big\{\mathsf{idx}\Big\}, \Big\{\top \Rightarrow \{\mathsf{ind}, \mathsf{ilg}\}\Big\}\Big\rangle.$$

*Finally, the* latex *operator is a bit more complicated. As a precondition it needs the presence of the* tex-*file and it produces as its effect* aux-, idx-, dvi, *and* log-*files unconditionally. In addition, we know that the citations will be correct if a* bbl-*file is present and that the index will be correct if an*

ind-*file is present:*

$$\begin{aligned}\mathsf{latex} = \Big\langle \Big\{\mathsf{tex}\Big\},\ &\Big\{\top \Rightarrow \{\mathsf{aux}, \mathsf{idx}, \mathsf{dvi}, \mathsf{log}\},\\ &\mathsf{bbl} \Rightarrow \mathsf{dvi_cite_ok},\\ &\neg\mathsf{bbl} \Rightarrow \neg\mathsf{dvi_cite_ok},\\ &\mathsf{ind} \Rightarrow \mathsf{dvi_ind_ok},\\ &\neg\mathsf{ind} \Rightarrow \neg\mathsf{dvi_ind_ok}\Big\}\Big\rangle.\end{aligned}$$

The semantics of operators is given by *state-transition functions*, i.e., mappings from states to states. Given a state $s$ and a set of postconditions *post*, $A(s, \mathit{post})$ denotes the **active effects** in $s$:

$$A(s, \mathit{post}) = \bigcup\{L \mid (\Gamma \Rightarrow L) \in \mathit{post}, s \models \Gamma\}.$$

The **state-transition function** $\theta_o$ induced by the operator $o$ is defined as follows:

$$\theta_o \colon \quad \mathbf{2}^{\Sigma} \rightarrow \mathbf{2}^{\Sigma}$$

$$\theta_o(s) = \begin{cases} s - \neg neg(A(s, \mathit{post}(o))) \cup pos(A(s, \mathit{post}(o))) & \text{if } s \models \mathit{pre}(o) \text{ and} \\ & A(s, \mathit{post}(o)) \not\models \bot \\ \text{undefined} & \text{otherwise} \end{cases}$$

In words, if the precondition of the operator is satisfied in state $s$ and the active effects are consistent, then state $s$ is mapped to the state $s'$ which differs from $s$ in that the truth values of *active effects* are forced to become true for positive effects and forced to become false for negative effects. If the precondition is not satisfied or the set of active effects is inconsistent, the result of the function is undefined.

In the planning formalism itself, we do not work on states but on *state specifications*. In general, this can lead to semantic problems. By restricting ourselves to state specifications that are sets of literals, however, the syntactic manipulations of the state specifications can be defined in a way such that they are *sound* in Lifschitz' (1986) sense.

Similarly to the active effects with respect to states, we define a corresponding function with respect to state specifications:

$$A(S, \mathit{post}) = \bigcup\{L \mid (\Gamma \Rightarrow L) \in \mathit{post}, S \models \Gamma\}.$$

Further, we define the **potentially active effects** as follows:

$$P(S, \mathit{post}) = \bigcup_{s \models S} A(s, \mathit{post}).$$

If for a state specification $S$ and an operator $o = \langle \mathit{pre}, \mathit{post}\rangle$, we have $A(S, \mathit{post}) \neq P(S, \mathit{post})$,[3] it means that the state specification resulting from the application of the state-transition functions might not be representable as a theory consisting of literals only. For this reason, we consider such an operator application as illegal, resulting in the illegal state specification $\bot$. We could be more liberal at this point and consider an operator application to a state specification only as illegal if the set of states resulting from applying the state-transition functions could definitely not be represented

3. Note that this can only happen if the state specification is incomplete.

as a theory consisting of literals only. Alternatively, we could consider all atoms mentioned in $P(S, \mathit{post}) - A(S, \mathit{post})$ as "unsafe" after the application of the operator and delete the literals $\neg(P(S, \mathit{post}) - A(S, \mathit{post}))$ from the state specification, but consider the resulting state specification still as "legal" if $P(S, \mathit{post})$ is consistent. Since there does not seem to exist a standard model for the execution of conditional effects in the presence of partial state specifications, we adopt the first alternative as one arbitrary choice. It should be noted, however, that this decision influences some of the results we present below.

Similarly to the rule that $A(S, \mathit{post}) \neq P(S, \mathit{post})$ leads to an illegal state specification, we require that if the precondition is not satisfied by all states in $\mathit{Mod}(S)$ or if the state specification is already inconsistent, the result of applying $o$ to $S$ results in $\bot$. This leads to the definition of the function $R$, which defines the outcome of applying an operator $o$ from the set of operators **O** to a state specification:

$$R\colon \quad \mathbf{2}^{\widehat{\Sigma}} \times \mathbf{O} \to \mathbf{2}^{\widehat{\Sigma}}$$

$$R(S, o) = \begin{cases} S - \neg A(S, \mathit{post}(o)) \cup A(S, \mathit{post}(o)) & \text{if } S \not\models \bot \text{ and} \\ & S \models \mathit{pre}(o) \text{ and} \\ & A(S, \mathit{post}(o)) \not\models \bot \text{ and} \\ & A(S, \mathit{post}(o)) = P(S, \mathit{post}(o)) \\ \bot & \text{otherwise} \end{cases}$$

**Example 2** *Using the propositional atoms and operators from Example 1, we assume the following two state specifications* $\mathsf{S}_1 = \{\mathsf{tex}, \mathsf{ind}\}$*, and* $\mathsf{S}_2 = \{\mathsf{tex}, \mathsf{ind}, \mathsf{bbl}, \mathsf{blg}\}$*. If we try to apply the operator* $\mathsf{latex}$ *to* $\mathsf{S}_1$*, we notice that this results in* $\bot$ *because*

$$\begin{aligned} A(\mathsf{S}_1, \mathit{post}(\mathsf{latex})) &= \{\mathsf{aux}, \mathsf{idx}, \mathsf{dvi}, \mathsf{log}, \mathsf{dvi_ind_ok}\}, \\ P(\mathsf{S}_1, \mathit{post}(\mathsf{latex})) &= A(\mathsf{S}_1, \mathit{post}(\mathsf{latex})) \cup \{\mathsf{dvi_cite_ok}, \neg\mathsf{dvi_cite_ok}\}, \end{aligned}$$

*i.e., we have* $A(\mathsf{S}_1, \mathit{post}(\mathsf{latex})) \neq P(\mathsf{S}_1, \mathit{post}(\mathsf{latex}))$*. On the other hand, we can apply* $\mathsf{bibtex}$ *successfully to* $\mathsf{S}_1$*:* $R(\mathsf{S}_1, \mathsf{bibtex}) = \mathsf{S}_2$.

It is easily verified that the syntactic operation on a state specification using the function $R$ corresponds to state transitions on the states described by the specification.

**Proposition 1** *Let $S$ be a state specification, $o$ be an operator, and $\theta_o$ be the induced state-transition function. If $R(S, o) \not\models \bot$, then*

$$\mathit{Mod}(R(S, o)) = \{s' \mid s' = \theta_o(s), s \models S\}.$$

*If $R(S, o) \models \bot$, then either*

1. *$\mathit{Mod}(S) = \emptyset$, or*
2. *there are two states $s_1, s_2 \in \mathit{Mod}(S)$ such that $A(s_1, \mathit{post}(o)) \neq A(s_2, \mathit{post}(o))$, or*
3. *there exists a state $s \in \mathit{Mod}(S)$ such that $\theta_o(s)$ is undefined.*

In other words, whenever $R(S, o)$ results in a "legal" specification, this specification describes the states that result from the application of the state-transition function $\theta_o$ to the states that satisfy the original state specification $S$. Further, if $R(S, o)$ is illegal, there are good reasons for it.

A **planning instance** is a tuple

$$\Pi = \langle \Xi, \mathbf{I}, \mathbf{G} \rangle,$$

where

- $\Xi = \langle \Sigma, \mathbf{O} \rangle$ is the **domain structure** consisting of a finite set of propositional atoms $\Sigma$ and a finite set of operators $\mathbf{O}$,
- $\mathbf{I} \subseteq \widehat{\Sigma}$ is the **initial state specification**, and
- $\mathbf{G} \subseteq \widehat{\Sigma}$ is the **goal specification**.[4]

When we talk about the **size of an instance**, symbolically $||\Pi||$, in the following, we mean the size of a (reasonable) encoding of the instance.

In the following, we use the notation $\mathbf{O}^\star$ to refer to the **set of finite sequences** of operators. Elements $\Delta$ of $\mathbf{O}^\star$ are called **plans**. Then $||\Delta||$ denotes the size of the plan, i.e., the number of operators in $\Delta$. We say that $\Delta$ is a $c$**-step plan** if $||\Delta|| \leq c$. The result of applying $\Delta$ to a state specification $S$ is recursively defined as follows:

$$\begin{aligned} Res: & \quad \mathbf{2}^{\widehat{\Sigma}} \times \mathbf{O}^\star \rightarrow \mathbf{2}^{\widehat{\Sigma}} \\ Res(S, \langle\rangle) &= S \\ Res(S, \langle o_1, o_2, \ldots, o_n \rangle) &= Res(R(S, o_1), \langle o_2, \ldots, o_n \rangle) \end{aligned}$$

A sequence of operators $\Delta$ is said to be a **plan for** $\Pi$ or a **solution of** $\Pi$ iff

1. $Res(\mathbf{I}, \Delta) \not\models \bot$ and
2. $Res(\mathbf{I}, \Delta) \models \mathbf{G}$.

**Example 3** *Let $\Sigma$ and $\mathbf{O}$ be the propositional atoms and operators introduced in Example 1 and consider the following planning instance:* $\Pi = \langle \langle \Sigma, \mathbf{O} \rangle, \{\mathsf{tex}, \mathsf{bib}, \neg\mathsf{ind}\}, \{\mathsf{dvi}, \mathsf{dvi_cite_ok}\} \rangle$. *In words, given a latex source file (*tex*) and a bibliography database (*bib*), we want to generate a dvi file (*dvi*) such that the citations in this file are correct (*dvi_cite_ok*). Furthermore, we do not know anything about the existence of a bbl-file or aux-file etc., but we know that there is no index file yet (*$\neg$ind*). The plan* $\Delta = \langle \mathsf{bibtex}, \mathsf{latex} \rangle$ *is a solution of* $\Pi$ *because the plan does not result in an illegal state specification and the resulting state specification entails* dvi *and* dvi_cite_ok.

Plans satisfying (1) and (2) above are "sound." In order to state this more precisely, we extend the notion of state transition functions for operators to state transition functions for plans. Let $\theta_\Delta$ be the state transition function corresponding to the composition of primitive state-transition functions induced by the operators in $\Delta = \langle o_1, \ldots, o_n \rangle$, i.e.,

$$\theta_{\langle o_1, \ldots, o_n \rangle} \quad = \quad \theta_{o_1} \circ \ldots \circ \theta_{o_n},$$

4. We could have been more liberal requiring that $\mathbf{G} \subseteq \mathcal{L}_\Sigma$. We have not done that in order to allow for a "fair" comparison with restricted planning formalisms.

such that $\theta_{\langle o_1,\ldots,o_n\rangle}(s)$ is defined iff $\theta_{\langle o_1,\ldots,o_i\rangle}(s)$ is defined for every $i$, $1 \leq i \leq n$. Using this notion, one can easily prove—using induction over the plan length—that any plan for an instance $\Pi$ is **sound** in Lifschitz' (1986) sense, i.e., corresponds to the application of state transition functions to the initial states.

**Proposition 2** *Let* $\Pi = \langle \Xi, \mathbf{I}, \mathbf{G} \rangle$ *be a planning instance and* $\Delta = \langle o_1, \ldots, o_n \rangle$ *be an element of* $\mathbf{O}^\star$. *If* $Res(\mathbf{I}, \Delta)$ *is consistent, then*

$$Mod(Res(\mathbf{I}, \Delta)) = \{s' \mid s' = \theta_\Delta(s), s \models \mathbf{I}\}.$$

*If* $Res(\mathbf{I}, \Delta)$ *is inconsistent, then either*

1. $Mod(\mathbf{I}) = \emptyset$, *or*

2. *there exists a (possibly empty) prefix* $\langle o_1, \ldots, o_i \rangle$ $(0 \leq i \leq n-1)$ *of* $\Delta$ *such that* $S = Res(\mathbf{I}, \langle o_1, \ldots, o_i \rangle)$ *and either*

   (a) *there are two states* $s_1, s_2 \in Mod(S)$ *such that* $A(s_1, post(o_{i+1})) \neq A(s_2, post(o_{i+1}))$, *or*

   (b) *there exists a state* $s \in Mod(S)$ *such that* $\theta_{o_{i+1}}(s)$ *is undefined.*

### 2.2 A Family of Propositional Planning Formalisms

The propositional variant of standard STRIPS (Fikes & Nilsson, 1971), which we will also call $\mathcal{S}$ in what follows, is a planning formalism that requires *complete state specifications, unconditional effects,* and *propositional atoms* as formulae in the precondition lists. Less restrictive planning formalisms can have the following additional features:

**Incomplete state specifications ($\mathcal{I}$):** The state specifications may not be complete.

**Conditional Effects ($\mathcal{C}$):** Effects can be conditional.

**Literals as formulae ($\mathcal{L}$):** The formulae in preconditions and effect conditions can be literals.

**Boolean formulae ($\mathcal{B}$):** The formulae in preconditions and effect conditions can be arbitrary boolean formulae.

These extensions can also be combined. We will use combinations of letters to refer to such multiple extensions. For instance, $\mathcal{S}_\mathcal{L}$ refers to the formalism $\mathcal{S}$ extended by literals in the precondition lists, $\mathcal{S}_{\mathcal{IC}}$ refers to the formalism allowing for incomplete state specifications and conditional effects, and $\mathcal{S}_{\mathcal{BIC}}$, finally, refers to the general planning formalism introduced in Section 2.1.

**Example 4** *When we consider the planning instance* $\Pi$ *from Example 3, it becomes quickly obvious that this instance has been expressed using* $\mathcal{S}_{\mathcal{LIC}}$. *The initial state specification is* incomplete, *the operator* `latex` *contains* conditional effects *and negative* literals *in some effect conditions. However, we do not need general* Boolean formulae *to express the instance.*

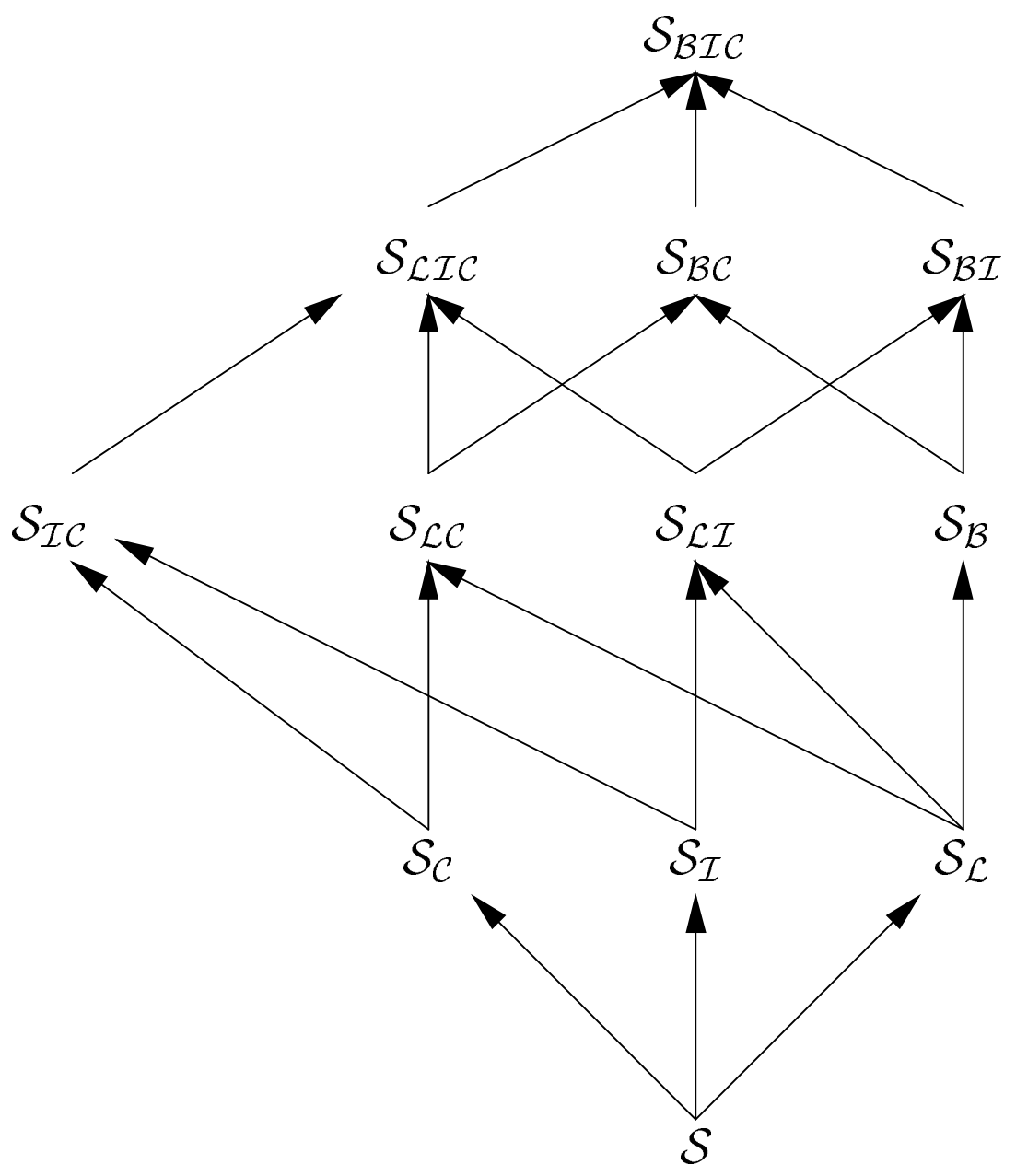

Figure 1: Planning formalisms partially ordered by syntactic restrictions

Figure 1 displays the partial order on propositional planning formalisms defined in this way. In the sequel we say that $\mathcal{X}$ **is a specialization** of $\mathcal{Y}$, written $\mathcal{X} \sqsubseteq \mathcal{Y}$, iff $\mathcal{X}$ is identical to $\mathcal{Y}$ or below $\mathcal{Y}$ in the diagram depicting the partial order.

Comparing this set of planning formalisms with the one Bäckström (1995) analyzed,[5] one notices that despite small differences in the presentation of the planning formalisms:

- $\mathcal{S}$ is the same as *common propositional strips* (CPS),
- $\mathcal{S}_{\mathcal{L}}$ is the same as *propositional strips with negative goals* (PSN), and
- $\mathcal{S}_{\mathcal{LI}}$ is the same as *ground Tweak* (GT).

### 2.3 The Computational Complexity of Planning in the $\mathcal{S}$-Family

While one would expect that planning in $\mathcal{S}$ is much easier than planning in $\mathcal{S}_{\mathcal{BIC}}$, it turns out that this is not the case, provided one takes a computational complexity perspective.

In analyzing the computational complexity of planning in different formalisms, we consider, as usual, the problem of deciding whether there *exists a plan* for a given instance—the **plan existence problem** (PLANEX). We will use a prefix referring to the planning formalism if we consider the existence problem in a particular planning formalism.

**Theorem 3** *$\mathcal{X}$-PLANEX is* PSPACE*-complete for all $\mathcal{X}$ with $\mathcal{S} \sqsubseteq \mathcal{X} \sqsubseteq \mathcal{S}_{\mathcal{BIC}}$.*

5. We do not consider planning formalisms identical to the SAS$^{+}$ formalism (Bäckström & Nebel, 1995), since we do not allow for multi-valued state variables.

**Proof.** PSPACE-hardness of $\mathcal{S}$-PLANEX follows from a result by Bylander (1994, Corollary 3.2).

Membership of $\mathcal{S}_{\mathcal{BIC}}$-PLANEX in PSPACE follows because we could, step by step, guess a sequence of operators, verifying at each step that the operator application leads to a legal follow up state specification and that the last operator application leads to a state specification that entails the goal specification. For each step, this verification can be carried out in polynomial space. The reason for this is that all the conditions in the definition of $R$ are verified by polynomially many calls to an NP-oracle. Therefore, $\mathcal{S}_{\mathcal{BIC}}$ can be decided on a non-deterministic machine in polynomial space, hence it is a member of PSPACE.

From that it follows that the plan existence problem for all formalisms that are in expressiveness between $\mathcal{S}$ and $\mathcal{S}_{\mathcal{BIC}}$—including both formalisms—is PSPACE-complete. ■

## 3. Expressiveness and Compilability between Planning Formalisms

Although there is no difference in the computational complexity between the formalisms in the $\mathcal{S}_{\mathcal{BIC}}$-family, there might nevertheless be a difference in how concisely planning domains and plans can be expressed. In order to investigate this question, we introduce the notion of *compiling planning formalisms.*

### 3.1 Compiling Planning Formalisms

As mentioned in the Introduction, we will consider a planning formalism $\mathcal{X}$ *as expressive as* another formalism $\mathcal{Y}$ if planning domains and plans formulated in formalism $\mathcal{Y}$ are *concisely expressible* in $\mathcal{X}$. We formalize this intuition by making use of what we call *compilation schemes*, which are *solution preserving mappings* with *polynomially sized results* from $\mathcal{Y}$ domain structures to $\mathcal{X}$ domain structures. While we restrict the size of the result of a compilation scheme, we do not require any bounds on the computational resources for the compilation. In fact, for measuring the *expressibility*, it is irrelevant whether the mapping is polynomial-time computable, exponential-time computable, or even non-recursive. At least, this seems to be the idea when the notion of *expressive power* is discussed in similar contexts (Baader, 1990; Erol et al., 1996; Gogic et al., 1995; Cadoli et al., 1996). If we want to use such compilation schemes in practice, they should be reasonably efficient, of course. However, if we want to prove that one formalism is *strictly more expressive* than another one, we have to prove that there is no compilation scheme regardless of how many computational resources such a compilation scheme might use.

So far, compilation schemes restrict only the size of domain structures. However, when measuring expressive power, the size of the generated plans should also play a role. In Bäckström's ESP-reductions (1995), the plan size must be identical. Similarly, the translation from $\mathcal{S}_{\mathcal{LC}}$ to $\mathcal{S}_{\mathcal{L}}$ proposed by Gazen and Knoblock (1997) seems to have as an implicit prerequisite that the plan length in the target formalism should be almost the same. When comparing the expressiveness of different planning formalisms, we might, however, be prepared to accept some growth of the plans in the target formalism. For instance, we may accept an additional constant number of operators, or we may even be satisfied if the plan in the target formalism is linearly or polynomially larger. This leads to the schematic picture of compilation schemes as displayed in Figure 2.

Although Figure 2 gives a good picture of the *compilation framework*, it is not completely accurate. First of all, a compilation scheme may introduce some auxiliary propositional atoms that are used to control the execution of newly introduced operators. These atoms should most likely have an initial value and may appear in the goal specification of planning instances in the target formalism. We will assume that the compilation scheme takes care of this and adds some literals to the initial state and goal specifications.

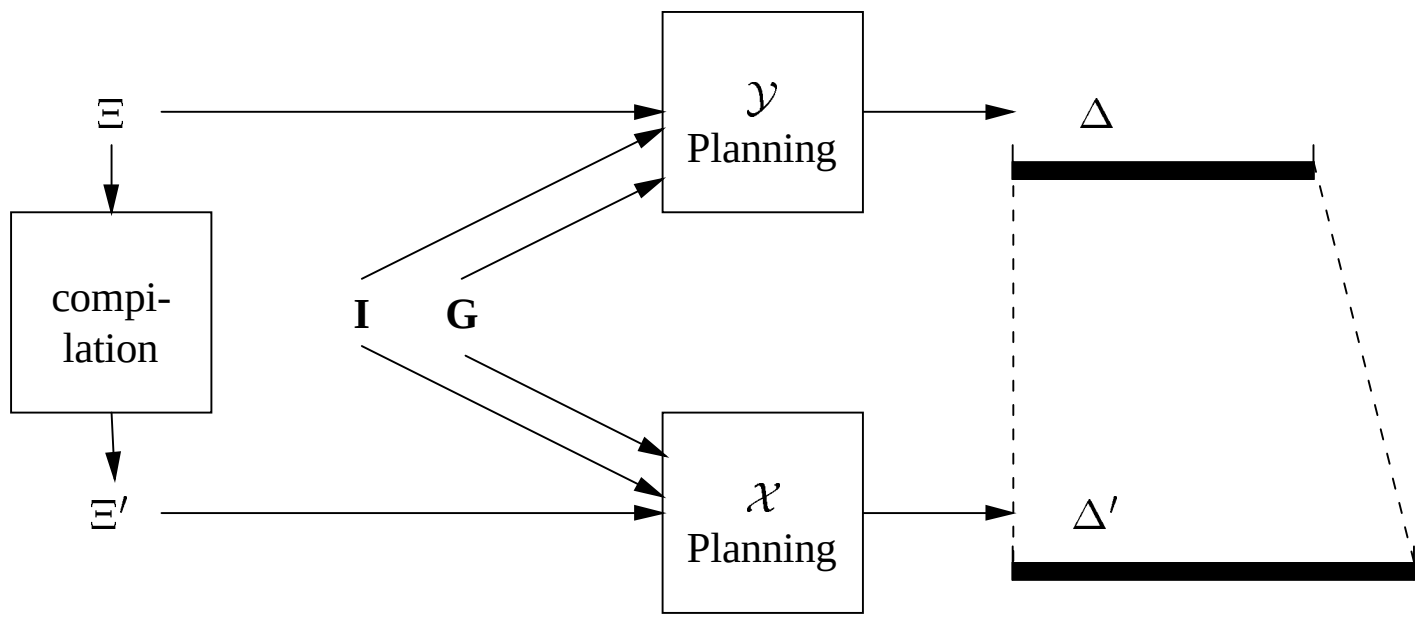

Figure 2: The compilation framework

Additionally, some translations of the initial state and goal specifications may be necessary. If we want to compile a formalism that permits for literals in preconditions and goals to one that requires atoms, some trivial translations are necessary. Similarly, if we want to compile a formalism that permits us to use partial state specification to a formalism that requires complete state specifications, a translation of the initial state specification is necessary. However, such *state translation functions* should be very limited. They should depend only on the set of symbols in the source formalism, should be "context-independent," i.e., the translation of a literal in a state specification should not depend on the whole specification, and they should be efficiently computable.

While the compilation framework is a theoretical tool to measure expressiveness, it has, of course, practical relevance. Let us assume that we have a reasonably fast planning system for a planning formalism $\mathcal{X}$ and we want to add a new feature to $\mathcal{X}$ resulting in formalism $\mathcal{Y}$. If we can come up with an *efficient* compilation scheme from $\mathcal{Y}$ to $\mathcal{X}$, this means we can easily integrate the new feature—either by using the compilation scheme or by modifying the planning algorithm minimally. If no compilation scheme exists, we probably would have problems integrating this feature. Finally, if only computationally expensive compilation schemes exist, we have an interesting situation. In this case, the off-line compilation costs may be high. However, since the compiled domain structure can be used for different initial and goal state specifications, the high off-line costs may be compensated by the efficiency gain resulting from using the $\mathcal{X}$ planning algorithm.[6] As it turns, however, this situation does not arise in analyzing compilability between the $\mathcal{S}_{\mathcal{BIC}}$ formalisms. Either we can identify a polynomial-time compilation scheme or we are able to prove that no compilation scheme exists.

6. This means that compilation schemes between planning formalisms are similar to knowledge compilations (Cadoli & Donini, 1997), where the fixed part of a computational problem is the domain structure and the variable part consists of the initial state and goal specifications. The main difference to the knowledge compilation framework is that we also take the (size of the) result into account. In other words we compile function problems instead of decision problems.

### 3.2 Compilation Schemes

Assume a tuple of functions $\mathbf{f} = \langle f_\xi, f_i, f_g, t_i, t_g \rangle$ that induce a function $F$ from $\mathcal{X}$-instances $\Pi = \langle \Xi, \mathbf{I}, \mathbf{G} \rangle$ to $\mathcal{Y}$-instances $F(\Pi)$ as follows:

$$F(\Pi) = \langle f_\xi(\Xi), f_i(\Xi) \cup t_i(\Sigma, \mathbf{I}), f_g(\Xi) \cup t_g(\Sigma, \mathbf{G}) \rangle.$$

If the following three conditions are satisfied, we call $\mathbf{f}$ a **compilation scheme from $\mathcal{X}$ to $\mathcal{Y}$**:

1. there exists a plan for $\Pi$ iff there exists a plan for $F(\Pi)$;

2. the **state-translation functions** $t_i$ and $t_g$ are **modular**, i.e., for $\Sigma = \Sigma_1 \cup \Sigma_2$, $S \subseteq \widehat{\Sigma}$, and $S \not\models \bot$, the functions $t_x$ (for $x = i, g$) satisfy

   $$t_x(\Sigma, S) = t_x(\Sigma_1, S \cap \widehat{\Sigma_1}) \cup t_x(\Sigma_2, S \cap \widehat{\Sigma_2}),$$

   and they are polynomial-time computable;

3. and the size of the results of $f_\xi$, $f_i$, and $f_g$ is polynomial in the size of the arguments.

Condition (1) states that the function $F$ induced by the compilation scheme $\mathbf{f}$ is solution-preserving. Condition (2) states requirements on the *on-line state-translation functions*. The result of these functions should be computable *element-wise*, provided the state specification is consistent. Considering the fact that these functions depend only on the original set of symbols and the state specification, this requirement does not seem to be very restrictive. Since the state-translation functions are on-line functions, we also require that the result should be efficiently computable.[7] Finally, condition (3) formalizes the idea that $\mathbf{f}$ is a compilation. For a *compilation* it is much more important that the result can be *concisely represented*, i.e., in polynomial space, than that the compilation process is fast. Nevertheless, we are also interested in *efficient compilation schemes*. We say that $\mathbf{f}$ is a **polynomial-time compilation scheme** if $f_\xi$, $f_i$, and $f_g$ are polynomial-time computable functions.

In addition to the resource requirements on the compilation process, we will distinguish between different compilation schemes according to the effects on the size of the plans solving the instance in the target formalism. If a compilation scheme $\mathbf{f}$ has the property that for every plan $\Delta$ solving an instance $\Pi$ there exists a plan $\Delta'$ solving $F(\Pi)$ such that $||\Delta'|| \leq ||\Delta|| + k$ for some positive integer constant $k$, $\mathbf{f}$ is a **compilation scheme preserving plan size exactly** (up to additive constants). If $||\Delta'|| \leq c \times ||\Delta|| + k$ for positive integer constants $c$ and $k$, then $\mathbf{f}$ is a **compilation scheme preserving plan size linearly**, and if $||\Delta'|| \leq p(||\Delta||, ||\Pi||)$ for some polynomial $p$, then $\mathbf{f}$ is a **compilation scheme preserving plan size polynomially**. More generally, we say that a planning formalism $\mathcal{X}$ is **compilable** to formalism $\mathcal{Y}$ (in polynomial time, preserving plan size exactly, linearly, or polynomially), if there exists a compilation scheme with the appropriate properties. We write $\mathcal{X} \preceq^x \mathcal{Y}$ in case $\mathcal{X}$ is compilable to $\mathcal{Y}$ or $\mathcal{X} \preceq^x_p \mathcal{Y}$ if the compilation can be done in polynomial time. The super-script $x$ can be $1$, $c$, or $p$ depending on whether the scheme preserves plan size exactly, linearly plan, or polynomially, respectively.

As is easy to see, all the notions of compilability introduced above are reflexive and transitive.

7. Although it is hard to imagine a modular state-translation function that is not polynomial time computable, some pathological function could, e.g., output translations that have exponential size in the *encoding* of the symbols.

**Proposition 4** *The relations $\preceq^x$ and $\preceq_p^x$ are transitive and reflexive.*

Furthermore, it is obvious that when moving upwards in the diagram displayed in Figure 1, there is always a polynomial-time compilation scheme preserving plan size exactly. If $\pi_i$ denotes the projection to the $i$-th argument and $\emptyset$ the function that returns always the empty set, the generic compilation scheme for moving upwards in the partial order is $\mathbf{f} = \langle \pi_1, \emptyset, \emptyset, \pi_2, \pi_2 \rangle$.

**Proposition 5** *If $\mathcal{X} \sqsubseteq \mathcal{Y}$, then $\mathcal{X} \preceq_p^1 \mathcal{Y}$.*

## 4. Compilability Preserving Plan Size Exactly

Proposition 5 leads to the question of whether there exist other compilation schemes than those implied by the specialization relation. Because of Proposition 5 and Proposition 4, we do not have to find compilation schemes for every pair of formalisms. It suffices to prove that $\mathcal{X}$ is compilable to $\mathcal{Y}$, in order to arrive at the conclusion that all formalisms that are below $\mathcal{X}$ are compilable to $\mathcal{Y}$ and formalisms above $\mathcal{Y}$.

A preview of the results of this section is given in Figure 3. We will establish two equivalence classes such that all members of each class are compilable to each other preserving plan size exactly. These two equivalence classes will be called $\mathcal{S}_{\mathcal{LI}}$- and $\mathcal{S}_{\mathcal{LIC}}$-class, in symbols $[\mathcal{S}_{\mathcal{LI}}]$ and $[\mathcal{S}_{\mathcal{LIC}}]$, naming them after their respective largest elements.

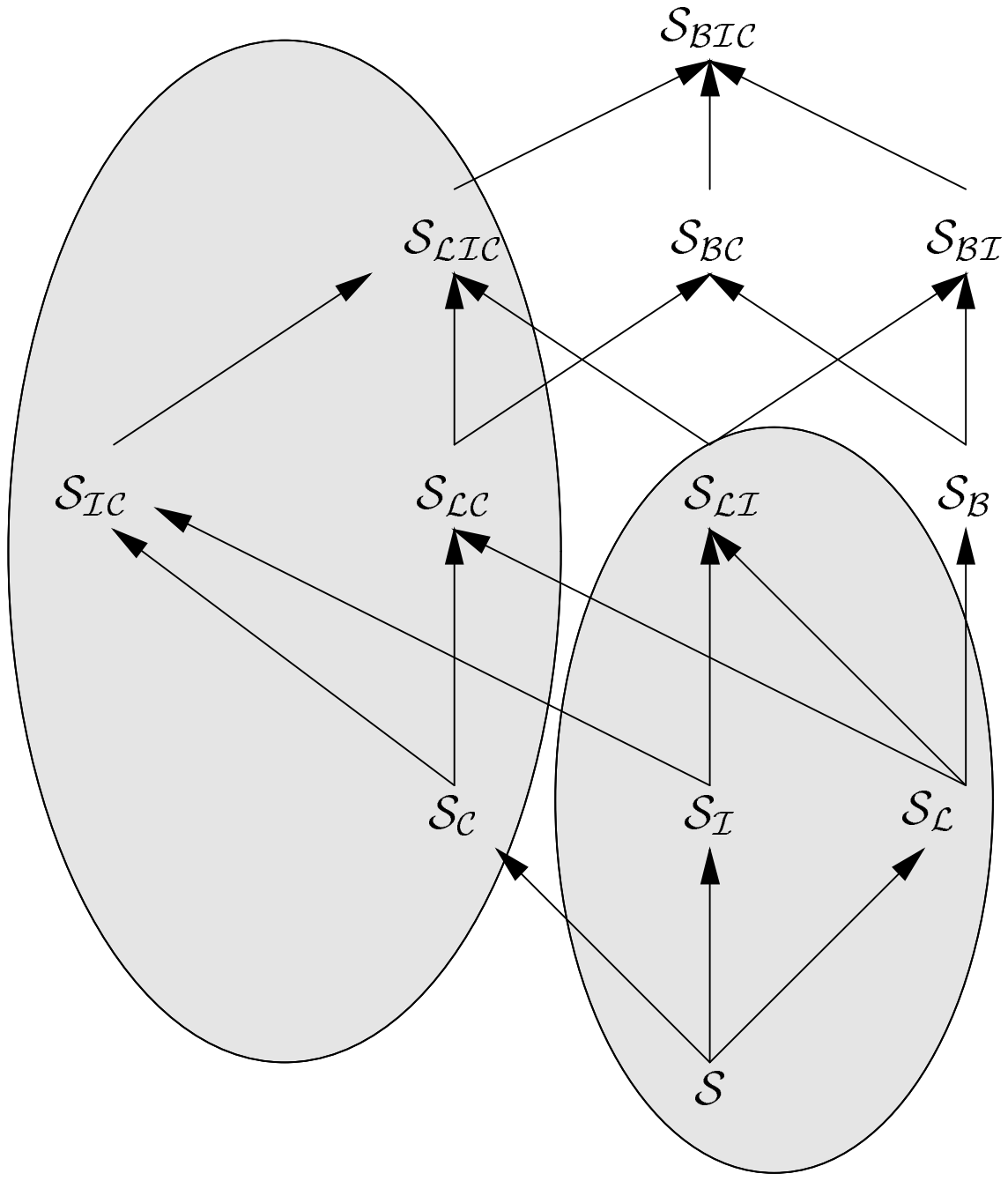

Figure 3: Equivalence classes of planning formalisms created by polynomial-time compilation schemes preserving plan size exactly

### 4.1 Planning Formalisms without Conditional Effects and Boolean Formulae

First, we will show that the formalisms analyzed by Bäckström (1995), namely, $\mathcal{S}_{\mathcal{LI}}$, $\mathcal{S}_{\mathcal{L}}$, and $\mathcal{S}$ are polynomial-time compilable into each other preserving plan size exactly. In fact, a fourth class can be added to this set, namely, $\mathcal{S}_{\mathcal{I}}$, which lies between $\mathcal{S}_{\mathcal{LI}}$ and $\mathcal{S}$.

In other words, using the notion of *compilability*, we get the same equivalence class as with Bäckström's ESP-reductions. Having a closer look at the proofs in Bäckström's (1995) paper reveals that this is not surprising at all because the ESP-reductions he used could be reformulated as compilation schemes. Since he used a quite different notation, we will nevertheless prove this claim from first principles.

The key idea in compiling planning formalisms with literals to formalisms that allow for atoms only is to consider $p$ and $\neg p$ as different atoms in the new formalism. For this purpose, we introduce $\widetilde{\Sigma} = \{\widetilde{p} \mid p \in \Sigma\}$, i.e., a disjoint copy of $\Sigma$. Further, if $L \subseteq \widehat{\Sigma}$, then $\sim L$ is a set where each negative literal $\neg p$ in $L$ is replaced by $\widetilde{p}$, i.e.,

$$\sim L = \begin{cases} \{p \in \Sigma \mid p \in L\} \cup \{\widetilde{p} \in \widetilde{\Sigma} \mid \neg p \in L\} & \text{if } L \not\models \bot, \\ \bot & \text{otherwise.} \end{cases}$$

Using $\Sigma \cup \widetilde{\Sigma}$ as the new set of atoms, one can translate state specifications and preconditions easily. In the postconditions we have to make sure that the intended semantics is taken care of, i.e., whenever $p$ is added, $\widetilde{p}$ must be deleted and *vice versa*.

Finally, we have to deal with the problem of partial state specifications. However, this not a problem when all effects are unconditional and the preconditions contain only atoms. In this case, we can safely assume that all atoms with unknown truth-value are false without changing the outcome of the application of an operator. Let $CWA_{\Sigma}(L)$ denote the **completion** of $L$ with respect to $\Sigma$, i.e.,

$$CWA_{\Sigma}(L) = \{\neg p \mid p \in \Sigma, p \notin L\} \cup L.$$

Using this function, we can transform a partial state specification into a complete specification without changing the outcome, i.e., we get the same plans.

**Theorem 6** *$\mathcal{S}_{\mathcal{LI}}$, $\mathcal{S}_{\mathcal{I}}$, $\mathcal{S}_{\mathcal{L}}$, and $\mathcal{S}$ are polynomial-time compilable to each other preserving plan size exactly.*

**Proof.** Since $\mathcal{S} \sqsubseteq \mathcal{S}_{\mathcal{I}} \sqsubseteq \mathcal{S}_{\mathcal{LI}}$ and $\mathcal{S} \sqsubseteq \mathcal{S}_{\mathcal{L}} \sqsubseteq \mathcal{S}_{\mathcal{LI}}$, it follows from Propositions 4 and 5 that we only have to show that $\mathcal{S}_{\mathcal{LI}} \preceq^1_p \mathcal{S}$ in order to prove the claim.

Let $\Pi = \langle \Xi, \mathbf{I}, \mathbf{G} \rangle$ be a $\mathcal{S}_{\mathcal{LI}}$-instance with $\Xi = \langle \Sigma, \mathbf{O} \rangle$. We translate each operator $o \in \mathbf{O}$ into the operator

$$\widetilde{o} = \langle \sim pre(o), \sim post(o) \cup \neg \sim \neg post(o) \rangle.$$

The set of all such operators is denoted by $\widetilde{\mathbf{O}}$. Now we can define the compilation scheme $\mathbf{f} = \langle f_\xi, f_i, f_g, t_i, t_g \rangle$ as follows:

$$\begin{aligned} f_\xi(\Xi) &= \langle \Sigma \cup \widetilde{\Sigma},\ \widetilde{\mathbf{O}} \rangle, \\ f_i(\Xi) &= \emptyset, \\ f_g(\Xi) &= \emptyset, \\ t_i(\Sigma, \mathbf{I}) &= CWA_{\Sigma \cup \widetilde{\Sigma}}(\sim \mathbf{I}), \\ t_g(\Sigma, \mathbf{G}) &= \sim \mathbf{G}. \end{aligned}$$

The scheme **f** obviously satisfies conditions (2) and (3), all the functions can be computed in polynomial time, and $F(\Pi)$ is a $\mathcal{S}$-instance.

Let $S \subseteq \widehat{\Sigma}$. Then it is obvious that

$$t_i(\Sigma, R(S, o)) = R(t_i(\Sigma, S), \widetilde{o}).$$

Let $\widetilde{\Delta} = \langle \widetilde{o_1}, \ldots, \widetilde{o_n} \rangle$ denote a sequence of operators corresponding to a sequence of operators $\Delta = \langle o_1, \ldots, o_n \rangle$. Using induction on plan length, it is easy to show that

$$\Delta \text{ is a plan for } \Pi \text{ iff } \widetilde{\Delta} \text{ is a plan for } F(\Pi),$$

i.e., condition (1) on compilation schemes is also satisfied. This means, **f** is in fact a compilation scheme. Further, since the plan size does not change, the compilation scheme preserves plan size exactly. Finally, because all functions in **f** can be computed in time polynomial in their arguments, **f** is a polynomial-time compilation scheme. ■

One view on this result is that it does not matter whether, from an expressivity point of view, we allow for atoms only or for literals and it does not matter whether we have complete or partial state specification—provided propositional formulae and conditional effects are not allowed.

### 4.2 Planning Formalisms with Conditional Effects but without Boolean Formulae

Interestingly, the view spelled out above generalizes to the case where conditional effects are allowed. Also in this case it does not matter whether only atoms or also literals are allowed and whether we have partial or complete state specifications. In proving that, however, there are two additional complications. Firstly, one must compile conditional effects over partial state specifications to conditional effects over complete state specifications. This is a problem because the condition $A(S, \mathit{post}(o)) = P(S, \mathit{post}(o))$ in the definition of the function $R$ must be tested. Secondly, when compiling a formalism with literals into a formalism that allows for atoms only, the condition $A(S, \mathit{post}(o)) \not\models \bot$ in the definition of $R$ must be taken care of. For this reason, we will prove this result in two steps.

As a first step, we show that $\mathcal{S}_{\mathcal{LIC}}$ can be compiled to $\mathcal{S}_{\mathcal{LC}}$. The problem in specifying such a compilation scheme is that the execution of an operator $o$ on a partial state specification leads to the illegal state if $A(S, \mathit{post}(o)) \neq P(S, \mathit{post}(o))$.

When considering our running example (Ex. 1), things are quite obvious. When a state specification does not contain the literal or the negation of the literal that is mentioned in the effect condition, then the illegal state specification results. For example, if a state specification does neither contain $\mathsf{bbl}$ nor $\neg\mathsf{bbl}$, then the result of executing $\mathsf{latex}$ is $\bot$. In the general case, however, things are less straightforward because effect literals can be produced by more than one conditional rule and an effect condition can consist of more than one literal.

Assuming without loss of generality (using a polynomial transformation) that the effects are all singleton sets, we have to check the following condition. Either one of the conditional effects with the same effect literal is activated—i.e., the effect condition is entailed by the partial state—or all of the conditional effects with the same effect literal are *blocked*, i.e., each effect condition contains a literal that is inconsistent with the state specification. If this is true, the original operator satisfies $A(S, \mathit{post}(o)) = P(S, \mathit{post}(o))$, otherwise the resulting state specification is inconsistent. For example, consider the following $\mathcal{S}_{\mathcal{LIC}}$ operator:

$$o_i = \langle \top, \{\{p, \neg q\} \Rightarrow \{\neg p\}, \{u, v\} \Rightarrow \{\neg p\}\} \rangle.$$

The application of this operator satisfies $A(S, \mathit{post}(o)) = P(S, \mathit{post}(o))$ iff either

1. $p$ and $\neg q$ are true in the state specification, or
2. $u$ and $v$ are true in the state specification, or
3. one of $p$ and $\neg q$ is false *and* one of $u$ and $v$ is false.

In all other cases, we get $A(S, \mathit{post}(o)) \neq P(S, \mathit{post}(o))$ and the result is the illegal state. In order to test for this condition in a formalism with complete states we introduce four new sets of atoms:

$$\begin{array}{rcl} \Sigma' & = & \{p' \mid p \in \Sigma\}, \\ \Sigma_+ & = & \{p_+ \mid p \in \Sigma\}, \\ \Sigma_- & = & \{p_- \mid p \in \Sigma\}, \\ \Upsilon & = & \{x_{i,j} \mid \text{for the } j\text{th conditional effect of } o_i\}. \end{array}$$

The atom $p'$ is true if either $p$ or $\neg p$ is part of the original partial state specification. The atom $p_+$ is set true by an operator if one of the conditional effects adds $p$ or if $p$ does not appear as an effect in the operator. The atom $p_-$ is set true by an operator if one of the conditional effects deletes $p$ or if $\neg p$ does not appear as an effect in the operator. Finally, atoms of the form $x_{i,j}$ are added by an action if the $j$th conditional effect in the $i$th operator is blocked by some effect condition. Using these new atoms, we could translate the above operator to

$$\begin{array}{rcl} \widetilde{o_i} & = & \langle \top, \{ \{p', q', p, \neg q\} \Rightarrow \{p', p_-, \neg p\}, \\ & & \quad \{u', v', u, v\} \Rightarrow \{p', p_-, \neg p\}, \\ & & \quad \{p', \neg p\} \Rightarrow \{x_{i,1}\}, \\ & & \quad \{q', q\} \Rightarrow \{x_{i,1}\}, \\ & & \quad \{u', \neg u\} \Rightarrow \{x_{i,2}\}, \\ & & \quad \{v', \neg v\} \Rightarrow \{x_{i,2}\}, \\ & & \quad \top \Rightarrow \Sigma_+ \cup \Sigma_- - \{p_-\}, \\ & & \quad \top \Rightarrow \{x_{m,j} \in \Upsilon \mid m \neq i\}\}\rangle. \end{array}$$

Let $\upsilon(i, j)$ be a function that returns $p_+$ or $p_-$, if $p$ or $\neg p$, respectively, is the effect of the $j$th conditional effect in the $i$th operator. Assuming now that the atoms from $\Sigma'$ are set according to their intended semantics and that the previous operator deleted all atoms from $\Sigma_+ \cup \Sigma_- \cup \Upsilon$, the following *test* operator checks whether the original operator would have led to an inconsistent result:

$$\mathit{test} \;=\; \left\langle \top, \left\{ \{\neg x_{i,j}, \neg \upsilon(i,j)\} \Rightarrow \bot \mid x_{i,j} \in \Upsilon \right\} \right\rangle.$$

Whenever we have $\neg x_{i,j}$, it means the $j$th conditional effect in the $i$th operator (which must be the previously executed operator) was not blocked. If in addition to that the effect of this conditional effect was not activated, i.e., $\neg \upsilon(i, j)$ is true, we would have $A(S, \mathit{post}(o)) \neq P(S, \mathit{post}(o))$ in the original formalism. For this reason, we force the illegal state. Conversely, if either $x_{i,j}$ is true for all $i$ and $j$ or if it is false for one $j$, but $\upsilon(i, j)$ is true, we would have $A(S, \mathit{post}(o)) = P(S, \mathit{post}(o))$ in the original formalism and do not need to force the illegal state.

We now could force, by using some extra literals, that after each operator $\widetilde{o_i}$ the *test* operator is applied. This would result in a compilation scheme that preserves plan size only *linearly*. However, it is possible to do better than that. The key idea is to merge the test operator for the $i$th step into the operator of step $i + 1$.

**Lemma 7** *$\mathcal{S}_{\mathcal{LIC}}$ is polynomial-time compilable to $\mathcal{S}_{\mathcal{LC}}$ preserving plan size exactly.*

**Proof.** Let $\Pi = \langle \Xi, \mathbf{I}, \mathbf{G} \rangle$ be a $\mathcal{S}_{\mathcal{LIC}}$-instance with $\Xi = \langle \Sigma, \mathbf{O} \rangle$. Without loss of generality, we assume that the postconditions of operators $o_i \in \mathbf{O}$ have the following form:

$$\mathit{post}(o_i) = \{L_{i,1} \Rightarrow l_{i,1}, \ldots, L_{i,m_i} \Rightarrow l_{i,m_i}\},$$

with $L_{i,j} \subseteq \widehat{\Sigma}$ and $l_{i,j} \in \widehat{\Sigma}$.

First, we introduce a number of new sets of symbols that are pairwise disjoint and disjoint from $\Sigma$:

$$\begin{aligned}
\Sigma' &= \{p' \mid p \in \Sigma\},\\
\Sigma^0_+ &= \{p^0_+ \mid p \in \Sigma\},\\
\Sigma^1_+ &= \{p^1_+ \mid p \in \Sigma\},\\
\Sigma^0_- &= \{p^0_- \mid p \in \Sigma\},\\
\Sigma^1_- &= \{p^1_- \mid p \in \Sigma\},\\
\Upsilon^0 &= \{x^0_{i,j} \mid \text{for the } j\text{th conditional effect of } o_i\},\\
\Upsilon^1 &= \{x^1_{i,j} \mid \text{for the } j\text{th conditional effect of } o_i\}.
\end{aligned}$$

For a given set of literals $L \subseteq \widehat{\Sigma}$, $L'$ denotes the set of primed literals, i.e., $L' = \{p' \mid p \in L\} \cup \{\neg p' \mid (\neg p) \in L\}$ and the function $s(\cdot)$ denotes the successor function modulo 2, i.e., $s(k) = (k+1) \bmod 2$. Further, the functions $v^k$ for $k = 0, 1$ shall be functions from $\Upsilon^k$ to $\Sigma^k_+ \cup \Sigma^k_-$ such that

$$v^k(i,j) = \begin{cases} q^k_+ \in \Sigma^k_+ & \text{if } (L_{i,j} \Rightarrow q) \in \mathit{post}(o_i),\\ q^k_- \in \Sigma^k_- & \text{if } (L_{i,j} \Rightarrow \neg q) \in \mathit{post}(o_i).\end{cases}$$

Now let $\mathit{post}^k(o_i)$ for $k = 0, 1$ be defined as follows

$$\begin{aligned}
\mathit{post}^k(o_i) = \{L \cup \sigma(L') \Rightarrow \{p, p', p^k_+\} \mid p \in \Sigma, (L \Rightarrow p) \in \mathit{post}(o_i)\} \cup\\
\{L \cup \sigma(L') \Rightarrow \{\neg p, p', p^k_-\} \mid p \in \Sigma, (L \Rightarrow \neg p) \in \mathit{post}(o_i)\},
\end{aligned}$$

let $\mathit{block}^k(o_i)$ for $k = 0, 1$ be defined as

$$\begin{aligned}
\mathit{block}^k(o_i) = \{\{q, q'\} \Rightarrow x^k_{i,j} \mid (L_{i,j} \Rightarrow l_{i,j}) \in \mathit{post}(o_i), (\neg q) \in L_{i,j}\} \cup\\
\{\{\neg q, q'\} \Rightarrow x^k_{i,j} \mid (L_{i,j} \Rightarrow l_{i,j}) \in \mathit{post}(o_i), q \in L_{i,j}\},
\end{aligned}$$

and let $\mathit{test}^k$ be defined as

$$\mathit{test}^k = \{\{\neg x^{s(k)}_{i,j}, \neg v^{s(k)}(i,j)\} \Rightarrow \bot \mid x^{s(k)}_{i,j} \in \Upsilon^{s(k)}\}.$$

Further, let $c^1$, $c^2$, and $g$ be fresh symbols not appearing in $\Sigma \cup \Sigma' \cup \Sigma^k_+ \cup \Sigma^k_- \cup \Upsilon^k$. Now we can define the pair of compiled operators $o^k_i$ $(k = 0, 1)$ corresponding to the original operator $o_i \in \mathbf{O}$:

$$\begin{aligned}
o^k_i = \langle \mathit{pre}(o_i) \cup \{c^k\}, \mathit{post}^k(o_i) \cup \mathit{block}^k(o_i) \cup \mathit{test}^k \cup\\
\{\top \Rightarrow \{\neg c^k, \neg g, c^{s(k)}\}\} \cup\\
\{\top \Rightarrow \Sigma^k_+ \cup \Sigma^k_- - \{v^k(i,j)\}\} \cup\\
\{\top \Rightarrow \{x^k_{m,j} \in \Upsilon^k \mid m \neq i\}\} \cup\\
\{\top \Rightarrow \neg\Sigma^{s(k)}_+ \cup \neg\Sigma^{s(k)}_- \cup \neg\Upsilon^{s(k)}\}.
\end{aligned}$$

This pair of compiled operators achieves the intended effects and keeps track of fully known atoms using $post^k$, checks which conditional effects are blocked using $block^k$, tests whether the execution of the previous operator satisfied the condition $A(S, post(o)) = P(S, post(o))$ using $test^k$, and setup the bookkeeping atoms for the next step. Using the atoms $c^k$, it is enforced that executing and testing is merged by parallelizing the test on step $i$ and execution of step $i+1$. In order to check the execution of the last step, we need an extra checking step:

$$o_g^k = \langle \{c^k\}, test^k \cup \{\top \Rightarrow \{g, \neg c^k\}\}\rangle.$$

Now we can specify a compilation scheme $\mathbf{f}$ from $\mathcal{S}_{\mathcal{LIC}}$ to $\mathcal{S}_{\mathcal{LC}}$ as follows:

$$\begin{array}{rcl} f_\xi(\Xi) & = & \langle \Sigma \cup \Sigma' \cup \Sigma_+^0 \cup \Sigma_+^1 \cup \Sigma_-^0 \cup \Sigma_-^1 \cup \Upsilon^0 \cup \Upsilon^1 \cup \{g, c^0, c^1\}, \\ & & \bigcup \{o_i^0, o_i^1 \mid o_i \in \mathbf{O}\} \cup \{o_g^0, o_g^1\}\rangle, \\ f_i(\Xi) & = & \{\neg g, c^0, \neg c^1\} \cup \neg\Sigma_+^0 \cup \neg\Sigma_+^1 \cup \neg\Sigma_-^0 \cup \neg\Sigma_-^1 \cup \neg\Upsilon^0 \cup \neg\Upsilon^1, \\ f_g(\Xi) & = & \{g\}, \\ t_i(\Sigma, \mathbf{I}) & = & CWA_\Sigma(\mathbf{I}) \cup CWA_{\Sigma'}(\{p' \mid p \in \Sigma, \{p, \neg p\} \cap \mathbf{I} \neq \emptyset\}), \\ t_g(\Sigma, \mathbf{G}) & = & \mathbf{G}. \end{array}$$

The scheme $\mathbf{f}$ obviously satisfies conditions (2), i.e., that the state-translation functions are modular, and (3), i.e., that the compilation functions have polynomially sized results. Further, all the functions can be computed in polynomial time, and $F(\Pi)$ is a $\mathcal{S}_{\mathcal{LC}}$-instance.

Assume $S \subseteq \widehat{\Sigma}$. Then it is obvious that

$$t_i(\Sigma, Res(S, \langle o_i \rangle)) \cap (\widehat{\Sigma} \cup \widehat{\Sigma'}) = Res(t_i(\Sigma, S) \cup f_i(\Xi), \langle o_i^0 \rangle) \cap (\widehat{\Sigma} \cup \widehat{\Sigma'}),$$

provided $Res(S, \langle o_i \rangle) \not\models \bot$. In case $Res(S, \langle o_i \rangle) \models \bot$, either $Res(t_i(\Sigma, S) \cup f_i(\Xi), \langle o_i^0 \rangle) \models \bot$ or $A(S, post(o_i)) \neq P(S, post(o_i))$. In the latter case, the application of any operator to $Res(t_i(\Sigma, S), \langle o_i^0 \rangle)$ leads to an inconsistent state because of the conditional effects in $test^1$, which is part of all postconditions of operators applicable in this state. Additionally, the same is true for the relation between $t_i(\Sigma, Res(S, \langle o_i, o_j \rangle))$ and $Res(t_i(\Sigma, S) \cup f_i(\Xi), \langle o_i^0, o_j^1 \rangle)$.

Let $\Delta' = \langle o_1', \ldots, o_n' \rangle$ denote a sequence of operators corresponding to a sequence of operators $\Delta = \langle o_1, \ldots, o_n \rangle$. Using induction on the plan length, it can be easily shown that

$$\Delta \text{ is a plan for } \Pi \text{ iff } \Delta'; o_g^k \text{ is a plan for } F(\Pi).$$

Further, since any plan solving the instance $F(\Pi)$ must have $o_g^k$ as the last operator, it follows that

$$\text{there exists a plan for } \Pi \text{ iff there exists a plan for } F(\Pi).$$

From that it follows immediately that $\mathbf{f}$ is a polynomial-time compilation scheme from $\mathcal{S}_{\mathcal{LIC}}$ to $\mathcal{S}_{\mathcal{LC}}$ preserving plan size exactly, which proves the claim. ■

Having proved that $\mathcal{S}_{\mathcal{LIC}}$ can be compiled to $\mathcal{S}_{\mathcal{LC}}$ preserving plan size exactly, it seems worth noting that this result depends on the semantics chosen for executing conditional operators on partial state specifications. For example, if we use an alternative semantics that deletes all the literals in $\neg(P(S, post(o)) - A(S, post(o))$ provided $P(S, post(o))$ is consistent, then there exists probably only a compilation scheme that preserves plan size linearly. If we use a semantics where the resulting state specification is legal when the application of all state-transformation functions leads to a theory that can be represented as a set of literals, it seems unlikely that there exists a scheme that preserves plan size polynomially. The reason for this pessimistic conjecture is that under this semantics it appears to be coNP-hard to determine whether the state specification resulting from applying a $\mathcal{S}_{\mathcal{LIC}}$-operator is legal.

As a second step in showing that partial state specifications and literals can be compiled away, we show that we can compile $\mathcal{S}_{\mathcal{LC}}$ to $\mathcal{S}_{\mathcal{C}}$. The key idea in the proof is the same as in the proof of Theorem 6. We replace each negative literal $\neg p$ by a new atom $\widetilde{p}$. In order to detect inconsistencies introduced by conditional effects, we add to each postcondition conditional effects of the form $\{p, \widetilde{p}\} \Rightarrow \bot$. Further, to check that the last operator in a plan does not introduce any inconsistencies, we force the application of a "checking" operator that contains the same conditional effects.

**Lemma 8** *$\mathcal{S}_{\mathcal{LC}}$ is polynomial-time compilable to $\mathcal{S}_{\mathcal{C}}$ preserving plan size exactly.*

**Proof.** Let $\Pi = \langle \Xi, \mathbf{I}, \mathbf{G} \rangle$ be a $\mathcal{S}_{\mathcal{LC}}$-instance with $\Xi = \langle \Sigma, \mathbf{O} \rangle$. Since $\Pi$ is a $\mathcal{S}_{\mathcal{LC}}$-instance, the postconditions of all operators $o \in \mathbf{O}$ have the following form:

$$\mathit{post}(o) = \{L_1 \Rightarrow K_1, \ldots, L_m \Rightarrow K_m\},$$

with $L_j, K_j \subseteq \widehat{\Sigma}$.

As in the proof of Theorem 6, $\widetilde{\Sigma}$ shall be a disjoint copy of $\Sigma$, and $\sim L$ is the set of atoms where each negative literal $\neg p$ is replaced by the atom $\widetilde{p}$. Now let $\widetilde{\mathit{post}(o)}$ be the following set

$$\widetilde{\mathit{post}(o)} = \{\sim L_j \Rightarrow (\sim K_j \cup \neg \sim \neg K_j) \mid (L_j \Rightarrow K_j) \in \mathit{post}(o)\}.$$

Further, let *cons* be the set of conditional effects

$$\mathit{cons} = \{\{p, \widetilde{p}\} \Rightarrow \bot \mid p \in \Sigma\},$$

let $g$ be an atom not appearing in $\Sigma$, let $\widetilde{o}$ be

$$\widetilde{o} = \langle \sim \mathit{pre}(o), \widetilde{\mathit{post}(o)} \cup \mathit{cons} \cup \{\top \Rightarrow \neg g\} \rangle,$$

let $\widetilde{\mathbf{O}} = \{\widetilde{o} \mid o \in \mathbf{O}\}$, and let the operator $o_g$ be

$$o_g = \langle \top, \mathit{cons} \cup \{\top \Rightarrow g\} \rangle.$$

Then we can specify a compilation scheme $\mathbf{f}$ from $\mathcal{S}_{\mathcal{LC}}$ to $\mathcal{S}_{\mathcal{C}}$ as follows:

$$\begin{aligned}
f_\xi(\Xi) &= \langle \Sigma \cup \widetilde{\Sigma} \cup \{g\},\ \widetilde{\mathbf{O}} \cup \{o_g\} \rangle, \\
f_i(\Xi) &= \{\neg g\}, \\
f_g(\Xi) &= \{g\}, \\
t_i(\Sigma, \mathbf{I}) &= \sim \mathbf{I} \cup \neg \sim \neg \mathbf{I}, \\
t_g(\Sigma, \mathbf{G}) &= \sim \mathbf{G}.
\end{aligned}$$

The scheme $\mathbf{f}$ obviously satisfies conditions (2) and (3), all the functions can be computed in polynomial time, and $F(\Pi)$ is a $\mathcal{S}_{\mathcal{C}}$-instance.

Assume $S \subseteq \widehat{\Sigma}$. Then it is obvious that

$$t_i(\Sigma, R(S, o)) = R(t_i(\Sigma, S), \widetilde{o}), \quad \text{provided } R(S, o) \not\models \bot.$$

In case $R(S, o) \models \bot$, either $R(t_i(\Sigma, S), \widetilde{o}) \models \bot$ or $\{p, \widetilde{p}\} \subseteq R(t_i(\Sigma, S), \widetilde{o})$ for some $p \in \Sigma$. In the latter case, the application of any operator to $R(t_i(\Sigma, S), \widetilde{o})$ leads to an inconsistent state because of the conditional effects in *cons*, which is part of all postconditions.

Let $\widetilde{\Delta} = \langle \widetilde{o_1}, \ldots, \widetilde{o_n} \rangle$ denote a sequence of operators corresponding to a sequence of operators $\Delta = \langle o_1, \ldots, o_n \rangle$. Using induction on the plan length, it can be easily shown that

$$\Delta \text{ is a plan for } \Pi \text{ iff } \widetilde{\Delta}; o_g \text{ is a plan for } F(\Pi).$$

Further, since any plan solving the instance $F(\Pi)$ must have $o_g$ as the last operator, it follows that

$$\text{there exists a plan for } \Pi \text{ iff there exists a plan for } F(\Pi).$$

It follows that **f** is polynomial-time compilation scheme from $\mathcal{S}_{\mathcal{LC}}$ to $\mathcal{S}_{\mathcal{C}}$ preserving plan size exactly, which proves the claim. ■

This result is, of course, not dependent on the semantics because both formalisms deal only with complete state specifications, and hence we always have $A(S, \mathit{post}(o)) = P(S, \mathit{post}(o))$.

**Theorem 9** *$\mathcal{S}_{\mathcal{LIC}}$, $\mathcal{S}_{\mathcal{LC}}$, $\mathcal{S}_{\mathcal{IC}}$, and $\mathcal{S}_{\mathcal{C}}$ are polynomial-time compilable to each other preserving plan size exactly.*

**Proof.** $\mathcal{S}_{\mathcal{LIC}} \preceq^1_p \mathcal{S}_{\mathcal{C}}$ follows from Lemma 8, Lemma 7 and Proposition 4. Using Propositions 4 and 5 and the fact that $\mathcal{S}_{\mathcal{C}} \sqsubseteq \mathcal{S}_{\mathcal{LC}} \sqsubseteq \mathcal{S}_{\mathcal{LIC}}$ and $\mathcal{S}_{\mathcal{C}} \sqsubseteq \mathcal{S}_{\mathcal{IC}} \sqsubseteq \mathcal{S}_{\mathcal{LIC}}$, the claim follows. ■

## 5. The Limits of Compilation when Preserving Plan Size Linearly

The interesting question is, of course, whether there are other compilation schemes preserving plan size exactly than those we have identified so far. As it turns out, this is not the case. We will prove that for all pairs of formalisms for which we have not identified a compilation scheme preserving plan size exactly, such a compilation scheme is impossible even if we allow for a linear increase of the plan size. For some pairs of formalisms we are even able to prove that a polynomial increase of the plan size would not help in establishing a compilation scheme. These results are, however, conditional based on an assumption that is slightly stronger than the $\mathsf{P} \neq \mathsf{NP}$ assumption. A preview of the results of this section is given in Table 1. The symbol $\sqsubseteq$ means that there exists a compilation scheme because the first formalism is a specialization of the second one. In all the other cases, we specify the separation and give the theorem number for this result.

### 5.1 Conditional Effects Cannot be Compiled Away

First of all, we will prove that conditional effects cannot be compiled away. The deeper reason for this is that with conditional effects, one can independently do a number of things in parallel, which is impossible in formalisms without conditional effects. If we consider, for example, the operator latex from Example 1, it is clear that it ”‘propagates”' the truth value of bbl and ind to dvi_cite_ok and dvi_ind_ok, respectively—provided the state specification satisfies the precondition.

| $\preceq^x$ | $\mathcal{S}_{\mathcal{BIC}}$ | $[\mathcal{S}_{\mathcal{LIC}}]$ | $\mathcal{S}_{\mathcal{BC}}$ | $\mathcal{S}_{\mathcal{BI}}$ | $\mathcal{S}_{\mathcal{B}}$ | $[\mathcal{S}_{\mathcal{LI}}]$ |
|---|---|---|---|---|---|---|
| $\mathcal{S}_{\mathcal{BIC}}$ | $=$ | $\not\preceq^p$ Cor. 15 | $\not\preceq^p$ Cor. 15 | $\not\preceq^c$ Cor. 12 | $\not\preceq^p$ Cor. 15 | $\not\preceq^p$ Cor. 15 |
| $[\mathcal{S}_{\mathcal{LIC}}]$ | $\sqsubseteq$ | $=$ | $\sqsubseteq$ | $\not\preceq^c$ **Theo. 11** | $\not\preceq^c$ Cor. 12 | $\not\preceq^c$ Cor. 12 |
| $\mathcal{S}_{\mathcal{BC}}$ | $\sqsubseteq$ | $\not\preceq^c$ Cor. 19 | $=$ | $\not\preceq^c$ Cor. 12 | $\not\preceq^c$ Cor. 12 | $\not\preceq^c$ Cor. 19 |
| $\mathcal{S}_{\mathcal{BI}}$ | $\sqsubseteq$ | $\not\preceq^p$ Cor. 15 | $\not\preceq^p$ **Theo. 14** | $=$ | $\not\preceq^p$ Cor. 15 | $\not\preceq^p$ Cor. 15 |
| $\mathcal{S}_{\mathcal{B}}$ | $\sqsubseteq$ | $\not\preceq^c$ **Theo. 18** | $\sqsubseteq$ | $\sqsubseteq$ | $=$ | $\not\preceq^c$ Cor. 19 |
| $[\mathcal{S}_{\mathcal{LI}}]$ | $\sqsubseteq$ | $\sqsubseteq$ | $\sqsubseteq$ | $\sqsubseteq$ | $\sqsubseteq$ | $=$ |

Table 1: Separation Results

It is obviously possible to come up with a set of exponentially many operators that can do the same thing in one step. However, it is unclear how to do that with less than exponentially many operators. In fact, we will show that this is impossible.

In order to illustrate this point, let us generalize the above example. We start with a set of $n$ propositional atoms $\Sigma_n = \{p_1, \ldots, p_n\}$ and a disjoint copy of this set: $\Sigma_n^{\#} = \{p_i^{\#} \mid p_i \in \Sigma_n\}$. Further, if $S \subseteq \widehat{\Sigma_n}$, then $S^{\#}$ shall denote the corresponding set of literals over $\widehat{\Sigma_n^{\#}}$, i.e.,

$$S^{\#} = \{p_i^{\#} \mid p_i \in S\} \cup \{\neg p_i^{\#} \mid \neg p_i \in S\}.$$

Consider now the following $\mathcal{S}_{\mathcal{LIC}}$ domain structure:

$$\begin{aligned}
\Sigma_{2n} &= \Sigma_n \cup \Sigma_n^{\#}, \\
\mathbf{O}_{2n} &= \Big\{\langle \top, \{p_i \Rightarrow p_i^{\#}, \neg p_i \Rightarrow \neg p_i^{\#} \mid p_i \in \Sigma_n\}\rangle\Big\}, \\
\Xi_{2n} &= \langle \Sigma_{2n},\ \mathbf{O}_{2n}\rangle.
\end{aligned}$$

From the construction it follows that for all pairs $(\mathbf{I}, \mathbf{G})$ such that $\mathbf{I}$ is a consistent and complete set over $\widehat{\Sigma_n}$ and $\mathbf{G} \subseteq \mathbf{I}^{\#}$, the instance $\Pi = \langle \Xi_{2n}, \mathbf{I}, \mathbf{G}\rangle$ has a one-step plan. Conversely, for all pairs $(\mathbf{I}, \mathbf{G})$ with $\mathbf{G} \cap \widehat{\Sigma^{\#}} \not\subseteq \mathbf{I}^{\#}$, there does not exist a solution.

Trying to define a $\mathcal{S}_{\mathcal{BI}}$ domain structure polynomially sized in $||\Xi_{2n}||$ with the same property seems to be impossible, even if we allow for $c$-step plans. However, in trying to prove this, it turns out that an additional condition on the state-translation function is needed.

We say that the state-translation functions are **local** iff for all state specifications $S$ and for $\Sigma_1 \cap \Sigma_2 = \emptyset$ we have

$$t_i(\Sigma_1, S \cap \widehat{\Sigma_1}) \cap t_g(\Sigma_2, S \cap \widehat{\Sigma_2}) = \emptyset.$$

With locality as an additional condition on state-translation functions we could easily prove that conditional effects cannot be compiled away. Instead of doing so we will show, however, that it is possible to derive a weaker condition from the definition of compilation schemes that will be enough to prove the impossibility result. This weaker condition is *quasi-locality* of state-translation functions relative to a given set of symbols $\Sigma$, which in turn is based on the notion of *universal literals*. A literal $l$ is called a **universal literal** for given state-translation functions on $\Sigma$ iff one of the following conditions is satisfied:

1. for all $p \in \Sigma$: $l \in t_i(\{p\}, \{p\})$, or
2. for all $p \in \Sigma$: $l \in t_i(\{p\}, \{\neg p\})$, or
3. for all $p \in \Sigma$: $l \in t_i(\{p\}, \emptyset)$, or
4. for all $p \in \Sigma$: $l \in t_g(\{p\}, \{p\})$, or
5. for all $p \in \Sigma$: $l \in t_g(\{p\}, \{\neg p\})$, or
6. for all $p \in \Sigma$: $l \in t_g(\{p\}, \emptyset)$.

Let $\mathbf{U}$ denote the set of universal literals. Now we define **quasi-locality** of state-translation functions relative to a set of propositional atoms $\Sigma$ and the induced set of universal literals $\mathbf{U}$ as follows. For each $S \subseteq \widehat{\Sigma}$ such that $S \not\models \bot$ and for all pairs $\Sigma_1, \Sigma_2 \subseteq \Sigma$ with $\Sigma_1 \cap \Sigma_2 = \emptyset$, we have

$$t_i(\Sigma_1, S \cap \widehat{\Sigma_1}) \cap t_g(\Sigma_2, S \cap \widehat{\Sigma_2}) \subseteq \mathbf{U}.$$

In words, the only *non-local* literals in quasi-local state-translation functions are the universal literals.

**Lemma 10** *For a given compilation scheme $\mathbf{f} = \langle f_\xi, f_i, f_g, t_i, t_g \rangle$ and natural number $n$, there exists a set of atoms $\Sigma \subseteq \mathbf{\Sigma}$ such that $|\Sigma| \geq n$ and $t_i$ and $t_g$ are quasi-local on $\Sigma$.*

**Proof.** Let $t\colon \mathbf{\Sigma} \to \mathbf{2^\Sigma}$ be a function that has as the result the union of all results for all possible translations of a literal returned by the state-translation functions, i.e.,

$$\begin{aligned} t(p) = {} & t_i(\{p\}, \{p\}) \cup t_i(\{p\}, \{\neg p\}) \cup t_i(\{p\}, \emptyset) \cup \\ & t_g(\{p\}, \{p\}) \cup t_g(\{p\}, \{\neg p\}) \cup t_g(\{p\}, \emptyset). \end{aligned}$$

Set $\mathbf{S} = \mathbf{\Sigma}$ and $\mathbf{U} = \emptyset$. Now we choose an infinite subset $\mathbf{S}'$ of $\mathbf{S}$ such that either

1. for all $p \in \mathbf{S}'$, there are only finitely many other atoms $q \in \mathbf{S}'$ such that $(t(p) \cap t(q)) - \mathbf{U} \neq \emptyset$, or if such an infinite subset of $\mathbf{S}$ does not exist,
2. $\mathbf{S}'$ has a universal literal $l \notin \mathbf{U}$ and we set $\mathbf{U}' = \mathbf{U} \cup \{l\}$.

   Note that such an infinite subset $\mathbf{S}'$ must exist. The reason is that some literal $l \notin \mathbf{U}$ must occur for infinitely many atoms in $t$ over $\mathbf{S}$ because we could not find an infinite subset satisfying condition (1). Because for a single atom there are only six possible ways to generate $l$, there must exist an infinite subset such that this literal occurs in all of either $t_x(\{p\}, \{p\})$, $t_x(\{p\}, \{\neg p\})$, or $t_x(\{p\}, \emptyset)$ (for $x = i, g$) and in this subset $l$ is a universal literal.

If we can pick a subset satisfying the first condition, we can choose from it a finite subset $\Sigma$ with any desired cardinality such that the state-translation functions are quasi-local with respect to $\Sigma$ and $\mathbf{U}$.

Otherwise we repeat the selection process with $\mathbf{S}'$ and $\mathbf{U}'$ until condition (1) is satisfied. This selection process can only be repeated finitely often because otherwise there are some atoms $p$ such that $t(p)$ has an infinite result, which is impossible because the state-translation functions are polynomial-time computable and can therefore have only finite results.

This demonstrates that there always *exists* a set of propositional atoms such that the state-translation functions are quasi-local. However, we might not be able to effectively determine this set. ■

Using this result, we are finally able to prove the non-existence of compilation schemes for compiling conditional effects away when preserving plan size linearly.

**Theorem 11** *$\mathcal{S}_{\mathcal{LIC}}$ cannot be compiled to $\mathcal{S}_{\mathcal{BI}}$ preserving plan size linearly.*

**Proof.** Assume for contradiction that there exists a compilation scheme $\mathbf{f}$ from $\mathcal{S}_{\mathcal{LIC}}$ to $\mathcal{S}_{\mathcal{BI}}$ preserving plan size linearly, which compiles the domain structure $\Xi_{2n}$ defined above into the $\mathcal{S}_{\mathcal{BI}}$ domain structure

$$f_\xi(\Xi_{2n}) = \Xi'_{2n} = \langle \Sigma'_{2n}, \mathbf{O}'_{2n} \rangle.$$

Because of Lemma 10 we can assume that the set of atoms $\Sigma_{2n}$ is chosen such that the translation functions are quasi-local on this set.

Let us now consider all initial state specifications $\mathbf{I}$ that are consistent and complete over $\Sigma_n$ and do not contain only positive or only negative literals:

$$\mathbf{I} \in \mathbf{2}^{\widehat{\Sigma_n}} - \{\Sigma_n, \neg\Sigma_n\}.$$

Obviously, there are $2^n - 2$ such state specifications. By assumption, each $\mathcal{S}_{\mathcal{BI}}$ instance of the following form

$$\langle \Xi'_{2n},\ t_i(\Sigma_{2n}, \mathbf{I}) \cup f_i(\Xi_{2n}),\ t_g(\Sigma_n, \mathbf{I}^{\#}) \cup f_g(\Xi_{2n}) \rangle$$

has a $c$-step plan. Since there are only $O(|\mathbf{O}'_{2n}|^c)$ different $c$-step plans, which is a number polynomial in the size of $\Xi_{2n}$, the same plan $\Delta$ is used for different initial states—provided $n$ is sufficiently large.

Suppose that the plan $\Delta$ is used for the pairs $(\mathbf{I}'_1, \mathbf{G}'_1), (\mathbf{I}'_2, \mathbf{G}'_2)$, which result from $\mathbf{I}_1$ and $\mathbf{I}_2$:

$$\begin{aligned}
\mathbf{I}'_1 &= t_i(\Sigma_n, \mathbf{I}_1) \cup f_i(\Xi_{2n}) \\
\mathbf{G}'_1 &= t_g(\Sigma_n, \mathbf{I}_1^{\#}) \cup f_g(\Xi_{2n}) \\
\mathbf{I}'_2 &= t_i(\Sigma_n, \mathbf{I}_2) \cup f_i(\Xi_{2n}) \\
\mathbf{G}'_2 &= t_g(\Sigma_n, \mathbf{I}_2^{\#}) \cup f_g(\Xi_{2n})
\end{aligned}$$

Since $\mathbf{I}_1 \neq \mathbf{I}_2$, $\mathbf{I}_1$ and $\mathbf{I}_2$ must differ on at least one atom, say $p$. Without loss of generality we assume $p \in \mathbf{I}_1$ and $\neg p \in \mathbf{I}_2$. Since $\Delta$ is a successful plan from $\mathbf{I}'_1$ to $\mathbf{G}'_1$ and because $t_g$ is modular, it follows that

$$Res(\mathbf{I}'_1, \Delta) \supseteq \mathbf{G}'_1 \supseteq t_g(\{p^{\#}\}, \{p^{\#}\}).$$

Some of the literals in $t_g(\{p^\#\},\{p^\#\})$ may be added by operators in $\Delta$ but none of the literals in $t_g(\{p^\#\},\{p^\#\})$ can be deleted by an operator in $\Delta$ without reestablishing this literal by another operator after its deletion. Because $\Delta$ contains only operators with unconditional effects, it adds and deletes the same literals regardless of the initial state.

Let us now assume that there exists a literal $l \in t_g(\{p^\#\},\{p^\#\})$ that is not added by $\Delta$. This implies that $l \in \mathbf{I}'_1$ and we have to distinguish three cases:

1. $l \in f_i(\Xi_{2n})$, from which we conclude that $l \in \mathbf{I}'_2$.
2. $l \in t_i(\{p^\#\},\emptyset) \subseteq \mathbf{I}'_1$, which also implies that $l \in \mathbf{I}'_2$.
3. $l \in t_i(\{q\},L)$ with $q \neq p^\#$ and $L \in \{\{q\},\{\neg q\},\emptyset\}$. Because we assumed that the state-translation functions are quasi-local on $\Sigma_{2n}$, $l$ must be a universal literal. If $l$ is universal for $t_i$, then we will have $l \in \mathbf{I}'_2$ because the possible initial states contain positive and negative literals as well as no literal for some elements from $\Sigma_{2n}$. If $l$ is universal for $t_g$, it is present in $\mathbf{G}'_1$ and in $\mathbf{G}'_2$ for the same reason. Further, because $l$ is not added by $\Delta$ and $\Delta$ is a valid plan from $\mathbf{I}'_2$ to $\mathbf{G}'_2$, it must also be part of of $\mathbf{I}'_2$.

In other words, all literals $l \in t_g(\{p^\#\},\{p^\#\})$ that are not added by $\Delta$ are already in $\mathbf{I}'_1$ and $\mathbf{I}'_2$. From that we conclude that

$$Res(\mathbf{I}'_2,\Delta) \supseteq t_g(\{p^\#\},\{p^\#\}).$$

Now let

$$\begin{aligned}
\mathbf{G}''_2 &= t_g(\Sigma_{2n}-\{p^\#\},\mathbf{I}^\#_2-\{\neg p^\#\}) \cup f_g(\Xi_{2n}),\\
\mathbf{G}'''_2 &= \mathbf{G}''_2 \cup t_g(\{p^\#\},\{p^\#\})\\
&= t_g(\Sigma_{2n},\mathbf{I}^\#_2-\{\neg p^\#\}\cup\{p^\#\}) \cup f_g(\Xi_{2n}).
\end{aligned}$$

Because $t_g$ is modular, it is clear that $\mathbf{G}'_2 \supseteq \mathbf{G}''_2$ and therefore $Res(\mathbf{I}'_2,\Delta) \supseteq \mathbf{G}''_2$. Because $\Delta$ achieves $\mathbf{G}''_2$ as well as $t_g(\{p^\#\},\{p^\#\})$, it follows that (again because $t_g$ is modular), $\Delta$ achieves also $\mathbf{G}'''_2$.

Since $\langle \Xi_{2n},\mathbf{I}_2,\mathbf{I}^\#_2-\{\neg p^\#\}\cup\{p^\#\}\rangle$ does not have any plan, there should not be any plan for $\langle \Xi'_{2n},\mathbf{I}'_2,\mathbf{G}'''_2\rangle$. The fact that $\Delta$ is a plan for this instance implies that $\mathbf{f}$ cannot be a compilation scheme, which is the desired contradiction. ■

Using Propositions 4 and 5 as well as Theorem 9, this result can be generalized as follows (see also Table 1).

**Corollary 12** *$\mathcal{S}_{\mathcal{BIC}}$, $\mathcal{S}_{\mathcal{BC}}$, and $[\mathcal{S}_{\mathcal{LIC}}]$ cannot be compiled to $\mathcal{S}_{\mathcal{BI}}$ or any formalism specializing $\mathcal{S}_{\mathcal{BI}}$ preserving plan size linearly.*

This answers the question of whether more space efficient compilation schemes from $\mathcal{S}_{\mathcal{LC}}$ to $\mathcal{S}$ than the one proposed by Gazen and Knoblock (1997) are possible. Even assuming unbounded computational resources for the compilation process, a more space efficient compilation scheme is impossible—provided that the compilation should preserve plan size linearly.[8] If we allow polynomially larger plans, then efficient compilation schemes are possible (see Section 6).

8. This result demonstrates that the choice of the semantics can be very important. If we interpret conditional effects sequentially as Brewka and Hertzberg (1993) do, then there exists an straightforward compilation scheme preserving plan size exactly.

## 5.2 Non-Uniform Complexity Classes

In the next section we make use of so-called *non-uniform complexity classes*, which are defined using *advice-taking machines*, in order to prove the impossibility of a compilation scheme. An **advice-taking Turing machine** is a Turing machine with an **advice oracle**, which is a (not necessarily recursive) function $a$ from positive integers to bit strings. On input $I$, the machine loads the bit string $a(||I||)$ and then continues as usual. Note that the oracle derives its bit string only from the length of the input and not from the contents of the input. An advice is said to be **polynomial** if the oracle string is polynomially bounded by the instance size. Further, if $\mathsf{X}$ is a complexity class defined in terms of resource-bounded machines, e.g., $\mathsf{P}$ or $\mathsf{NP}$, then $\mathsf{X}/\mathsf{poly}$ (also called **non-uniform** $\mathsf{X}$) is the class of problems that can be decided on machines with the same resource bounds and polynomial advice.

Because of the advice oracle, the class P/poly appears to be much more powerful than P. However, it seems unlikely that P/poly contains all of NP. In fact, one can prove that $\mathsf{NP} \subseteq \mathsf{P}/\mathsf{poly}$ implies certain relationships between uniform complexity classes that are believed to be very unlikely. For stating this result, we first have to introduce the *polynomial hierarchy*.

Let $\mathsf{X}$ be a class of decision problems. Then $\mathsf{P}^{\mathsf{X}}$ denotes the class of decision problems $P$ that can be decided in polynomial time by a deterministic Turing machine that is allowed to use a procedure—a so-called **oracle**—for deciding a problem $Q \in \mathsf{X}$, whereby executing the procedure does only cost constant time. Similarly, $\mathsf{NP}^{\mathsf{X}}$ denotes the class of decision problems $P$ such that there is a nondeterministic Turing-machine that solves all instances of $P$ in polynomial time using an oracle for $Q \in \mathsf{X}$. Based on these notions, the sets $\Delta_k^p$, $\Sigma_k^p$, and $\Pi_k^p$ are defined as follows:[9]

$$\begin{aligned}
\Delta_0^p &= \Sigma_0^p = \Pi_0^p = \mathsf{P},\\
\Delta_{k+1}^p &= \mathsf{P}^{\Sigma_k^p},\\
\Sigma_{k+1}^p &= \mathsf{NP}^{\Sigma_k^p},\\
\Pi_{k+1}^p &= \mathsf{coNP}^{\Sigma_k^p}.
\end{aligned}$$

Thus, $\Sigma_1^p = \mathsf{NP}$ and $\Pi_1^p = \mathsf{coNP}$. The set of all classes defined in this way is called the **polynomial hierarchy**, denoted by PH. Note that

$$\mathsf{PH} = \bigcup_{k\geq 0} \Delta_k^p = \bigcup_{k\geq 0} \Sigma_k^p = \bigcup_{k\geq 0} \Pi_k^p \subseteq \mathsf{PSPACE}.$$

Further we have, $\Delta_k^p \subseteq \Sigma_k^p \cap \Pi_k^p$ and $\Sigma_k^p, \Pi_k^p \subseteq \Delta_{k+1}$. As with other classes, it is unknown whether the inclusions between the classes are proper. However, it is strongly believed that this is the case, i.e., that the hierarchy is truly infinite.

Based on the firm belief that the polynomial hierarchy is proper, the above mentioned question of whether $\mathsf{NP} \subseteq \mathsf{P}/\mathsf{poly}$ can be answered. It has been shown that $\mathsf{NP} \subseteq \mathsf{P}/\mathsf{poly}$ would imply that the *polynomial hierarchy* collapses on the second level (Karp & Lipton, 1982), i.e., $\Sigma_2^p = \Pi_2^p$. This, however, is considered to be quite unlikely. Further, it has been shown that $\mathsf{NP} \subseteq \mathsf{coNP}/\mathsf{poly}$ or $\mathsf{coNP} \subseteq \mathsf{NP}/\mathsf{poly}$ implies that the polynomial hierarchy collapses at the third level (Yap, 1983), i.e., $\Sigma_3^p = \Pi_3^p$, which again is considered to be very unlikely. We will use these result for proving that for some pairs of formalisms it is very unlikely that one formalism can be compiled into the other one.

9. The super-script $p$ is only used to distinguish these sets from the analogous sets in the Kleene hierarchy.

### 5.3 On the Expressive Power of Partial State Specifications and Boolean Formulae

In all the cases considered so far, operators over partial state specifications could be compiled to operators over complete state specifications, i.e., partial state specifications did not add any expressiveness. This is no longer true, however, if we also allow for arbitrary boolean formulae in preconditions and effect conditions. In this case, we can decide the coNP-complete problem of whether a formula is a tautology by deciding whether a one-step plan exists. Asking, for example, if the $\mathcal{S}_{\mathcal{BI}}$-instance $\langle \Sigma, \{\langle \varphi, g \rangle\}, \emptyset, \{g\}\rangle$ has a plan is equivalent to asking whether $\varphi$ is a tautology.

Let the **one-step plan existence problem** (1-PLANEX) be the PLANEX problem restricted to plans of size one. From the above it is evident that $\mathcal{S}_{\mathcal{BIC}}$-1-PLANEX and $\mathcal{S}_{\mathcal{BI}}$-1-PLANEX are coNP-hard. Let $p$ be some fixed polynomial, then the **polynomial step plan-existence problem** ($p$-PLANEX) is the PLANEX problem restricted to plans that have length bounded by $p(n)$, if $n$ is the size of the planning instance. As is easy to see, this problem is in NP for all formalisms except $\mathcal{S}_{\mathcal{BIC}}$ and $\mathcal{S}_{\mathcal{BI}}$. The reason is that after guessing a sequence of operators and state specifications of polynomial size, one can verify for each step in polynomial time that the precondition is satisfied by the current state specification and produces the next state specification. Since there are only polynomially many steps, the overall verification takes only polynomial time.

**Proposition 13** *$\mathcal{X}$-$p$-PLANEX can be solved in polynomial time on a nondeterministic Turing machine for all formalisms different from $\mathcal{S}_{\mathcal{BIC}}$ and $\mathcal{S}_{\mathcal{BI}}$.*

From the fact that $\mathcal{S}_{\mathcal{BI}}$-1-PLANEX is coNP-hard and, e.g., $\mathcal{S}_{\mathcal{BC}}$-p-PLANEX is in NP, it follows almost immediately that there is no *polynomial-time* compilation scheme from $\mathcal{S}_{\mathcal{BI}}$ to $\mathcal{S}_{\mathcal{BC}}$ that preserves plan length polynomially (if NP $\neq$ coNP). However, even if we allow for unbounded computational resources of the compilation process, a proof technique first used by Kautz and Selman (1992) can be used to show that such a compilation scheme cannot exist (provided $\Sigma_3^p \neq \Pi_3^p$).

**Theorem 14** *$\mathcal{S}_{\mathcal{BI}}$ cannot be compiled to $\mathcal{S}_{\mathcal{BC}}$ preserving plan size polynomially, unless $\Sigma_3^p = \Pi_3^p$.*

**Proof.** Let $\varphi$ be a propositional formula of size $n$ in conjunctive normal form with three literals per clause. As a first step, we construct for each $n$ a $\mathcal{S}_{\mathcal{BI}}$ domain structure $\Xi_n$ with size polynomial in $n$ and the following properties. Unsatisfiability of an arbitrary 3CNF formula $\varphi$ of size $n$ is equivalent to 1-step plan existence for the $\mathcal{S}_{\mathcal{BI}}$-1-PLANEX instance $\langle \Xi_n, \mathbf{I}_\varphi, \{g\}\rangle$, where $\mathbf{I}_\varphi$ can be computed in polynomial time from $\varphi$.

Given a set of $n$ atoms, denoted by $P_n$, we define the set of clauses $\mathbf{A}_n$ to be the set containing all clauses with three literals that can be built using these atoms. The size of $\mathbf{A}_n$ is $O(n^3)$, i.e., polynomial in $n$. Let $\mathbf{D}_n$ be a set of new atoms $p_{\{l_1,l_2,l_3\}}$ corresponding one-to-one to the clauses in $\mathbf{A}_n$. Further, let

$$\Phi_n = \bigwedge \Big\{ (l_1 \vee l_2 \vee l_3 \vee p_{\{l_1,l_2,l_3\}}) \mid \{l_1,l_2,l_3\} \in \mathbf{A}_n \Big\}.$$

We now construct a $\mathcal{S}_{\mathcal{BI}}$ domain structure $\Xi_n = \langle \Sigma_n, \mathbf{O}_n \rangle$ for all formulae of size $n$ as follows:

$$\begin{aligned} \Sigma_n &= P_n \cup \mathbf{D}_n \cup \{g\}, \\ \mathbf{O}_n &= \{\langle \{\neg \Phi_n\}, \{g\}\rangle\}. \end{aligned}$$

Let $\mathbf{C}$ be a function that determines for all 3CNF formulae $\varphi$, which atoms in $\mathbf{D}_n$ correspond to the clauses in the formula , i.e.,

$$\mathbf{C}(\varphi) = \{p_{\{l_1,l_2,l_3\}} \mid \{l_1, l_2, l_3\} \in \varphi\}.$$

Now, the initial state for any particular formula $\varphi$ of size $n$ is computed as follows:

$$\mathbf{I}_\varphi = \neg\mathbf{C}(\varphi) \cup (\mathbf{D}_n - \mathbf{C}(\varphi)) \cup \{\neg g\}.$$

From the construction, it follows that there exists a one-step plan for $\langle\Sigma_n, \mathbf{O}_n, \mathbf{I}_\varphi, \{g\}\rangle$ iff $\varphi$ is unsatisfiable.

Let us now assume that there exists a compilation scheme $\mathbf{f}$ from $\mathcal{S}_{\mathcal{BI}}$ to $\mathcal{S}_{\mathcal{BC}}$ preserving plan size polynomially. Further, let us assume that the $\mathcal{S}_{\mathcal{BI}}$ domain structure $\Xi_n$ is compiled to the $\mathcal{S}_{\mathcal{BC}}$ domain structure $\Xi'_n = \langle\Sigma'_n, \mathbf{O}'_n\rangle$. Using this compiled domain structure, we can construct the following advice-taking Turing machine.

On input of a formula $\varphi$ of size $n$, we load the advice $\langle\Xi'_n, f_i(\Sigma_n, \mathbf{O}_n), f_g(\Sigma_n, \mathbf{O}_n)\rangle$. This advice is polynomial because $\Xi_n$ is polynomial in the size of $\varphi$ and a compilation scheme generates only polynomially larger domain structures. Because $t_i$ is a polynomial-time function and $\mathbf{I}_\varphi$ can be computed from $\varphi$ in polynomial time, we can compute

$$\mathbf{I}' = t_i(\Sigma_n, \mathbf{I}_\varphi) \cup f_i(\Sigma_n, \mathbf{O}_n)$$

in polynomial time. Also the goal specification

$$\mathbf{G}' = t_g(\Sigma_n, \{g\}) \cup f_g(\Sigma_n, \mathbf{O}_n)$$

can be computed in polynomial time. Finally, we decide the $p$-PLANEX problem on the resulting $\mathcal{S}_{\mathcal{BC}}$-instance $\langle\Xi'_n, \mathbf{I}', \mathbf{G}'\rangle$. From Proposition 13 we know that this can be done in polynomial time on a nondeterministic Turing machine.

Because deciding $p$-PLANEX for $\langle\Xi'_n, \mathbf{I}', \mathbf{G}'\rangle$ is equivalent to deciding 1-PLANEX for $\langle\Xi_n, \mathbf{I}_\varphi, \{g\}\rangle$, which is in turn equivalent to deciding unsatisfiability of $\varphi$, it follows that we can decide a coNP-complete problem on a nondeterministic, polynomial advice-taking Turing machine in polynomial time. From that it follows that coNP $\subseteq$ NP/poly. Using Yap's (1983) result, the claim follows. ■

Using Proposition 4 and Proposition 5, the above result generalizes as follows (see also Table 1).

**Corollary 15** *$\mathcal{S}_{\mathcal{BIC}}$ and $\mathcal{S}_{\mathcal{BI}}$ cannot be compiled to any of the other planning formalisms preserving plan size polynomially, unless $\Sigma_3^p = \Pi_3^p$.*

If we restrict the form of the formulae, however, we may be able to devise compilation schemes from $\mathcal{S}_{\mathcal{BI}}$ to, e.g., $\mathcal{S}_{\mathcal{B}}$. Reconsidering the proof of the last theorem, it turns out that it is essential to use the negation of a CNF formula as a precondition. If we restrict ourselves to CNF formulae in preconditions, it seems possible to move from partial to complete state descriptions using ideas similar to the ones used in the proof of Lemma 7.

However, no such compilation scheme will work for $\mathcal{S}_{\mathcal{BIC}}$. The reason is the condition $A(S, post(o)) = P(S, post(o))$ in the definition of the function $R$. If this condition is not satisfied, the result of the operator is inconsistent. This condition could be easily employed to reduce *unsatisfiability* of CNF formulae to 1-step plan existence, which enables us to use the same technique as in the proof of the above theorem.

### 5.4 Circuit Complexity

For the next impossibility result we need the notions of *boolean circuits* and *families of circuits*. A **boolean circuit** is a directed, acyclic graph $C = (V, E)$, where the nodes $V$ are called **gates**. Each gate $v \in V$ has a type $type(v) \in \{\neg, \vee, \wedge, 1, 0\} \cup \{x_1, x_2, \ldots\}$. The gates with $type(v) \in \{1, 0, x_1, x_2, \ldots\}$ have in-degree zero, the gates with $type(v) \in \{\neg\}$ have in-degree one, and the gates with $type(v) \in \{\wedge, \vee\}$ have in-degree two. All gates except one have at least one outgoing edge. The gate with no outgoing edge is called the **output gate**. The gates with no incoming edges are called the **input gates**. The **depth** of a circuit is the length of the longest path from an input gate to the output gate. The **size** of a circuit is the number of gates in the circuit.

Given a *value assignment* to the variables $\{x_1, x_2, \ldots\}$, the circuit computes the value of the output gate in the obvious way. For example, for $x_1 = 1$ and $x_2 = 0$ we get the value 1 at the output gate of the circuit shown in Figure 4.

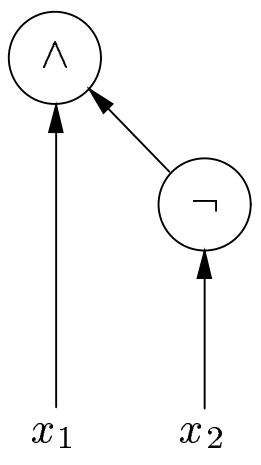

Figure 4: Example of a boolean circuit

Instead of using circuits for computing boolean functions, we can also use them for accepting words of length $n$ in $\{0, 1\}^\star$. A word $w = x_1 \ldots x_n \in \{0, 1\}^n$ is now interpreted as a value assignment to the $n$ input variables $x_1, \ldots, x_n$ of a circuit. The word is **accepted** iff the output gate has value 1 for this word. In order to deal with words of different length, we need one circuit for each possible length. A **family of circuits** is an infinite sequence $\mathbf{C} = (C_0, C_1, \ldots)$, where $C_n$ has $n$ input variables. The language accepted by such a family of circuits is the set of words $w$ such that $C_{||w||}$ accepts $w$.

Usually, one considers so-called **uniform** families of circuits, i.e., circuits that can be generated on a Turing machine with a $\log n$-space bound. Sometimes, however, also non-uniform families are interesting. For example, the class of languages accepted by non-uniform families of polynomially-sized circuits is just the class P/poly introduced in Section 5.2.

Using restrictions on the size and depth of the circuits, we can now define new complexity classes, which in their uniform variants are all subsets of P. One class that is important in the following is the class of languages accepted by uniform families of circuits with polynomial size and logarithmic depth, named $\mathsf{NC}^1$. Another class which proves to be important for us is defined in terms of non-standard circuits, namely circuits with gates that have *unbounded fan-in*. Instead of restricting the in-degree of each gate to be two at maximum, we now allow an unbounded in-degree. The class of languages accepted by families of polynomially sized circuits with unbounded fan-in and constant depth is called $\mathsf{AC}^0$.

From the definition, it follows almost immediately that $\mathsf{AC}^0 \subseteq \mathsf{NC}^1$. Moreover, it has been shown that there are some languages in $\mathsf{NC}^1$ that are not in the non-uniform variant of $\mathsf{AC}^0$, which implies that $\mathsf{AC}^0 \neq \mathsf{NC}^1$ (Furst, Saxe, & Sipser, 1984).

### 5.5 Boolean Formulae Cannot be Compiled to Conditional Effects

As we have seen in Section 5.3, Boolean formulae are quite expressive if they are used in combination with partial state specifications. However, what if all state specifications are complete? In this case, it seems to be possible to simulate the evaluation of CNF formulae by using conditional effects. In fact, it is possible to compile in polynomial-time, for example, $\mathcal{S}_{\mathcal{B}}$ to $\mathcal{S}_{\mathcal{LC}}$ preserving plan size linearly, provided all formulae are in conjunctive normal form. Each operator would have to be split into two operators, one that evaluates the clauses of all the formulae in the original operator and one that combines these evaluations and takes the appropriate actions, e.g., asserting $\perp$ if the precondition is not satisfied. Sequencing of these pairs of operators can be achieved by introducing some extra literals.

What can we say about the general case, however? When trying to simulate the evaluation of an arbitrary logical formula using conditional effects, it seems to be the case that we need as many operators as the nesting depth of the formula, which means that we would need plans that cannot be bounded to be only linearly longer than the original plans.

We will use the results sketched in Section 5.4 to separate $\mathcal{S}_{\mathcal{B}}$ and $\mathcal{S}_{\mathcal{LC}}$. In order to do so, let us view domain structures with fixed size plans as "machines" that accept languages. For all words $w$ consisting of $n$ bits, let

$$\Xi_n = \langle \Sigma_n \cup \{g\}, \mathbf{O}_n \rangle.$$

Assume that the atoms in $\Sigma_n$ are numbered from 1 to $n$. Then a word $w$ consisting of $n$ bits could be encoded by the set of literals

$$\mathbf{I}_w = \{p_i \mid \text{ if the } i\text{th bit of } w \text{ is } 1\} \cup \{\neg p_i \mid \text{ if the } i\text{th bit of } w \text{ is } 0\}.$$

Conversely, for a consistent state specification $S \in \widehat{\Sigma_n}$, let $w_S$ be a word such that the $i$th bit is 1 iff $p_i \in S$.

We now say that the $n$-bit word $w$ is **accepted with a one-step or $c$-step plan** by $\Xi_n$ iff there exists a one-step or $c$-step plan, respectively, for the instance

$$\Pi_n = \langle \langle \Sigma_n \cup \{g\}, \mathbf{O}_n \rangle, \mathbf{I}_w \cup \{\neg g\}, \{g\} \rangle.$$

Similarly to families of circuits, we also define families of domain structures, $\boldsymbol{\Xi} = (\Xi_0, \Xi_1, \ldots)$. The language accepted by such a family with a one-step (or $c$-step) plan is the set of words accepted using the domain structure $\Xi_n$ for words of length $n$. Borrowing the notion of uniformity as well, we say that a family of domain structures is **uniform** if it can be generated by a $\log n$-space Turing machine.

Papadimitriou has pointed out that the languages accepted by *uniform polynomially-sized boolean expressions* is identical to $\mathsf{NC}^1$ (Papadimitriou, 1994, p. 386). As is easy to see, a family of $\mathcal{S}_{\mathcal{B}}$ domain structures is nothing more than a family of boolean expressions, provided we use one-step plans for acceptance.

**Proposition 16** *The class of languages accepted by uniform families of $\mathcal{S}_{\mathcal{B}}$ domain structures using one-step plan acceptance is identical to* $\mathsf{NC}^1$.

If we now have a closer look at what the power of $c$-step plan acceptance for families of $\mathcal{S}_{\mathcal{LC}}$ domain structures is, it turns out that it is less powerful than $\mathsf{NC}^1$. In order to show that, we will first prove the following lemma that relates $c$-step $\mathcal{S}_{\mathcal{LC}}$ plans to circuits with gates of unbounded fan-in.

**Lemma 17** *Let $\Xi = \langle \Sigma, \mathbf{O} \rangle$ be a $\mathcal{S}_{\mathcal{LC}}$ domain structure, let $\mathbf{G} \subseteq \widehat{\Sigma}$, and let $\Delta$ be a $c$-step plan over $\Xi$. Then there exists a polynomially sized boolean circuit $C$ with unbounded fan-in and depth $7c+2$ such that $\Delta$ is a plan for $\langle \Xi, \mathbf{I}, \mathbf{G} \rangle$ iff the circuit $C$ has value 1 for the input $w_{\mathbf{I}}$.*

**Proof.** The general structure of a circuit for a $c$-step $\mathcal{S}_{\mathcal{LC}}$ plan is displayed in Figure 5. For each plan step (or level) $j$ and each atom $p_i$, there is a connection $p_i^j$. The connections on level 0 are the input gates, i.e., $p_i^0 = x_i$. The goal test is performed by an $\wedge$-gate that checks that all the goals are true on level $c$, in our case $\mathbf{G} = \{p_1, \neg p_2, p_n\}$. Further, using the $\vee$-gate, it is checked that no inconsistency was generated when executing the plan.

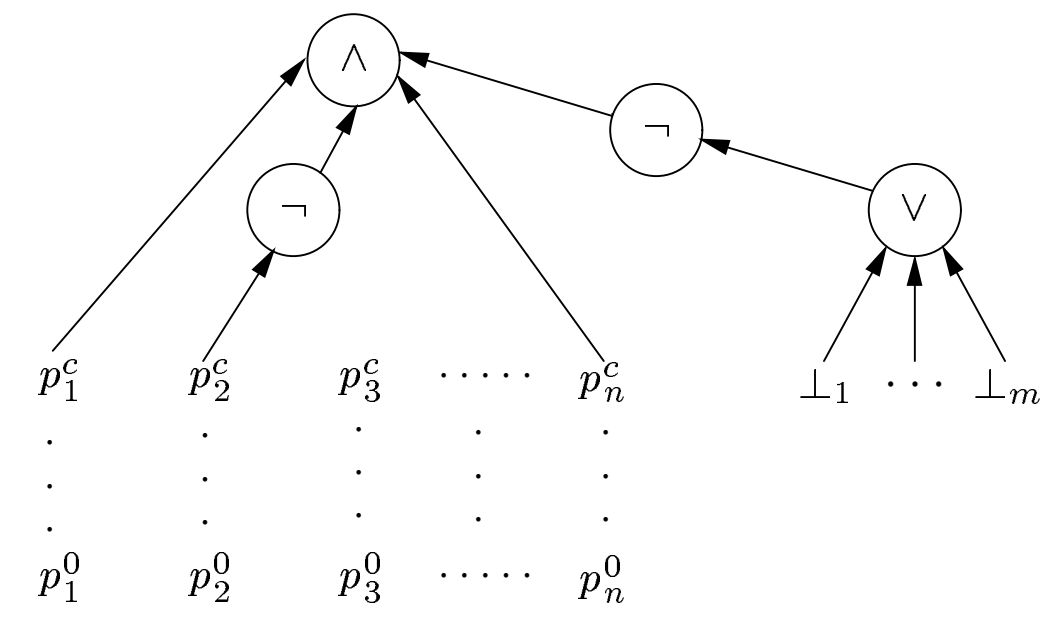

Figure 5: Circuit structure and goal testing for a $c$-step $\mathcal{S}_{\mathcal{LC}}$ plan

For each plan step $j$, it must be computed whether the precondition is satisfied and what the result of the conditional effects are. Figure 6 (a) displays the precondition test for the precondition $\{p_1, p_2, \neg p_3\}$. If the conjunction of the precondition literals is not true, $\bot_k$ becomes true, which is connected to the $\vee$-gate in Figure 5.

Without loss of generality (using a polynomial transformation), we assume that all conditional effects have the form $L \Rightarrow l$. Whether the effect $l$ is activated on level $j$ is computed by a circuit as displayed in Figure 6 (b), which shows the circuit for $\{p_1, \neg p_3\} \Rightarrow \neg p_i$.

Finally, all activated effects are combined by the circuit shown in Figure 6 (c). For all atoms $p_i$, we check whether both $p_i$ and $\neg p_i$ have been activated, which would set $\bot_r$ true. This is again one of the inputs of the $\vee$-gate in Figure 5. If neither $p_i$ nor $\neg p_i$ have been activated, the value of $p_i$ on level $j+1$ is determined by the value of $p_i$ on level $j$. Otherwise the value of $p_i$ on level $j+1$ is determined by the value of $p_i^{e,j}$, i.e., the activation value of the positive effect $p_i$ on level $j$.

The depths of the circuits in Figure 6 (b) and (c) dominate the depth of the circuit necessary to represent one plan step leading to the conclusion that a plan step can be represented using a circuit of depth 7. Adding the depth of the goal testing circuit, the claim follows. ■

The lemma implies that $\mathcal{S}_{\mathcal{LC}}$ $c$-step plan acceptance is indeed less powerful than $\mathcal{S}_{\mathcal{B}}$ 1-step plan acceptance, which means that a compilation scheme from $\mathcal{S}_{\mathcal{B}}$ to $\mathcal{S}_{\mathcal{LC}}$ preserving plan size linearly is impossible.

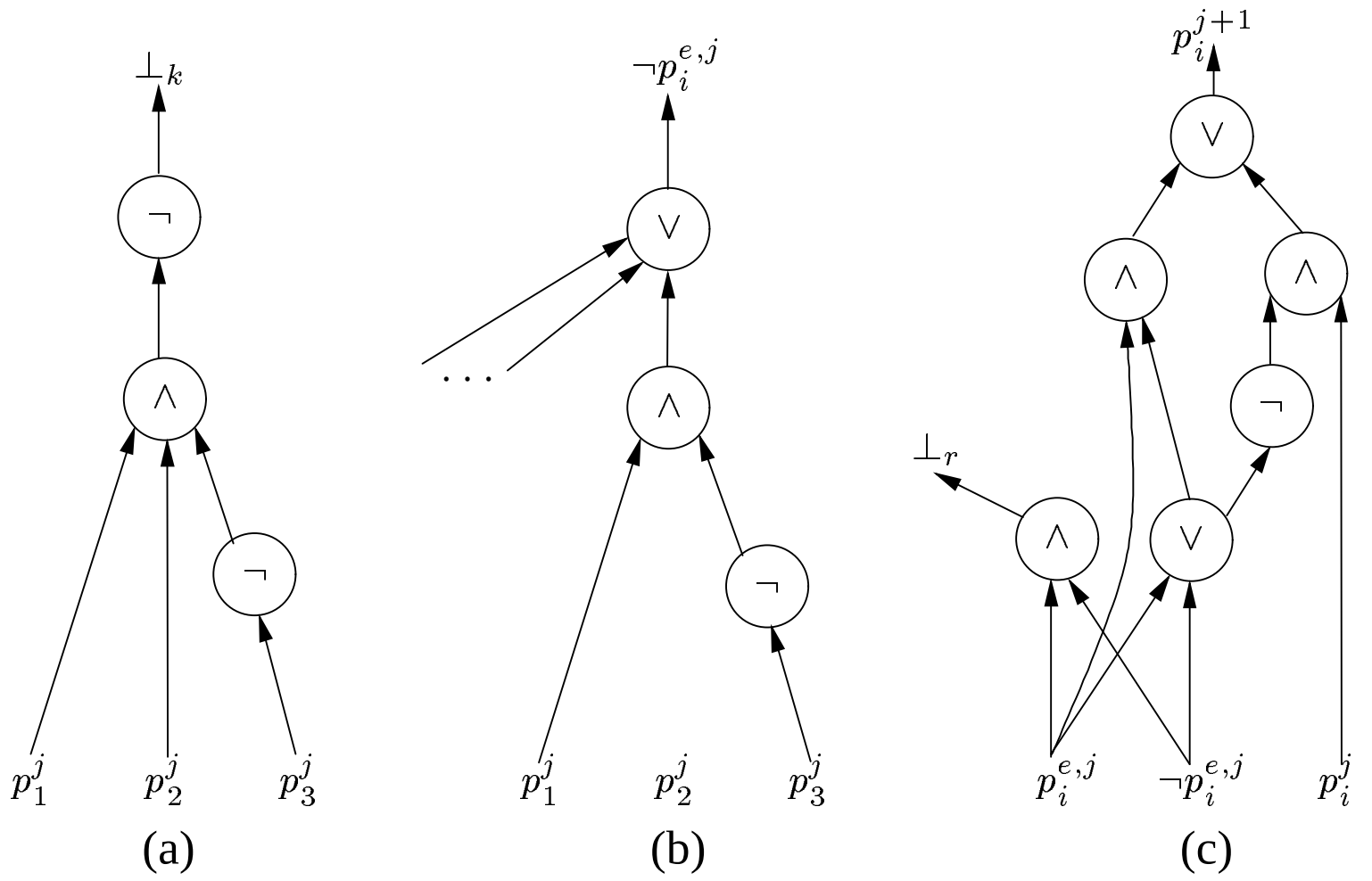

Figure 6: Circuit structure for precondition testing (a), conditional effects (b), and the computation of effects (c) for $\mathcal{S}_{\mathcal{LC}}$ operators

**Theorem 18** *$\mathcal{S}_{\mathcal{B}} \not\preceq^c \mathcal{X}$, for all members $\mathcal{X}$ of the $\mathcal{S}_{\mathcal{LIC}}$-class.*

**Proof.** We show that $\mathcal{S}_{\mathcal{B}} \not\preceq^c \mathcal{S}_{\mathcal{LC}}$, from which by Theorem 9 and Proposition 4 the claim follows.

Assume for contradiction that $\mathcal{S}_{\mathcal{B}} \preceq^c \mathcal{S}_{\mathcal{LC}}$. Let $\mathbf{\Xi} = (\Xi_0, \Xi_1, \ldots)$ be a uniform family of $\mathcal{S}_{\mathcal{B}}$ domain structures and $\mathbf{\Xi}' = (\Xi'_0, \Xi'_1, \ldots)$ be the $\mathcal{S}_{\mathcal{LC}}$ domain structures generated by a compilation scheme **f** that preserves plan size linearly. By Lemma 17 we know that for each $\mathcal{S}_{\mathcal{LC}}$ domain structure $\Xi'_n = \langle \Sigma'_n, \mathbf{O}'_n \rangle$ and given goal $\mathbf{G}'$ we can generate a polynomially sized, unbounded fan-in circuit with depth $7c + 2$ that tests whether a particular $c$-step plan achieves the goal. In order to decide $c$-step plan existence, we must test $O(|\mathbf{O}'_n|^c)$ different plans, which is polynomial in the size of $\Xi_n$ because **f** is a compilation scheme. For each plan, we can generate one test circuit, and by adding another $\vee$-gate we can decide $c$-step plan existence using a circuit with depth $7c+3$ and size polynomial in the size of $\Xi_n$. Further, since the state-translation functions are modular, the results of $t_i$ for fixed $\Sigma$ can be computed using an additional level of gates. Since by Proposition 16 all languages in $\mathsf{NC}^1$ are accepted by uniform families of $\mathcal{S}_{\mathcal{B}}$ domain structures using one-step plan acceptance, our assumption $\mathcal{S}_{\mathcal{B}} \preceq^c \mathcal{S}_{\mathcal{LC}}$ implies that we can accept all language in $\mathsf{NC}^1$ by (possibly non-uniform) $\mathsf{AC}^0$ circuits, which is impossible by the result of Furst and colleagues (1984). ■

Using the Propositions 4 and 5 again, we can generalize the above theorem as follows.

**Corollary 19** *$\mathcal{S}_{\mathcal{BC}}$ and $\mathcal{S}_{\mathcal{B}}$ cannot be compiled to $[\mathcal{S}_{\mathcal{LIC}}]$ or $[\mathcal{S}_{\mathcal{LI}}]$ preserving plan size linearly.*

## 6. Compilability Preserving Plan Size Polynomially

As has been shown in the previous section, only the compilation schemes induced by Propositions 4 and 5 and the ones identified in Section 4 allow for compilation schemes preserving plan size exactly. For all other pairs of formalisms we were able to rule out such compilation schemes—even if we allow linear growth of the resulting plans. Nevertheless, there might still be a chance for compilation schemes preserving plan size polynomially. Having shown that $\mathcal{S}_{\mathcal{BIC}}$ and $\mathcal{S}_{\mathcal{BI}}$ cannot be compiled to the other formalisms even if the plan can grow polynomially, we may still be able to find compilation schemes preserving plan size polynomially for the $\mathcal{S}_{\mathcal{BIC}}/\mathcal{S}_{\mathcal{BI}}$ pair and for the remaining formalisms.

A preview of the results of this section is given in Figure 7. As it can be seen, we are able

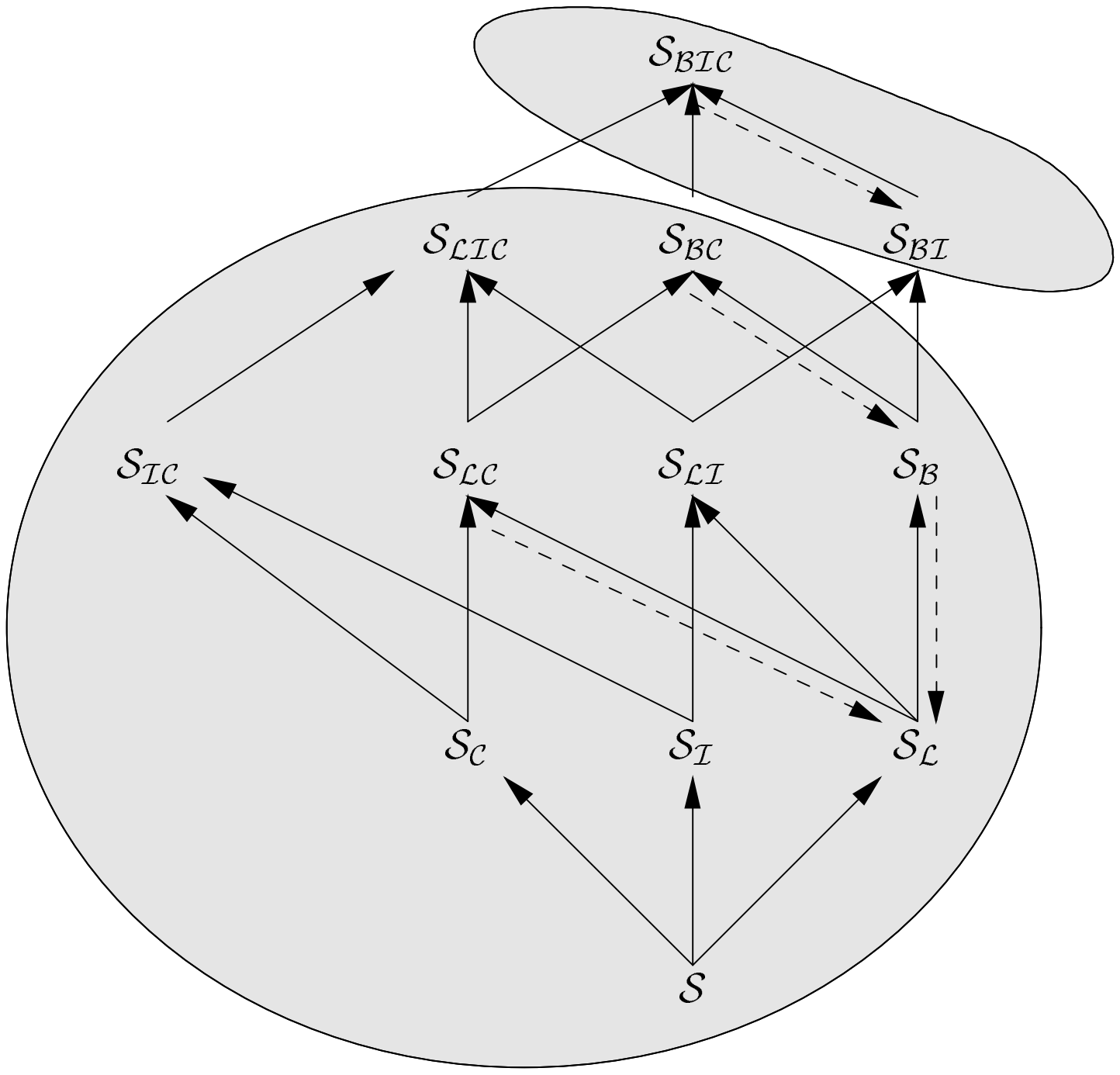

Figure 7: Equivalence classes of planning formalisms created by polynomial-time compilation schemes preserving plan size polynomially. Compilation schemes constructed in this section are indicated by dashed lines

to establish compilation schemes preserving plan size polynomially for all pairs of formalisms for which we have not proved the impossibility of such compilation schemes.

### 6.1 Compiling Conditional Effects Away for Partial State Specifications

The first compilation scheme we will develop is one from $\mathcal{S}_{\mathcal{BIC}}$ to $\mathcal{S}_{\mathcal{BI}}$. As before, we assume that the conditional effects have only singleton effect sets. Further, since we can use arbitrary boolean formulae in the effect conditions in $\mathcal{S}_{\mathcal{BIC}}$, we assume that there is only one rule for each effect literal. Using a simple polynomial transformation, arbitrary sets of operators can be brought into this form. This simplifies checking the condition $A(S, post(o)) = P(S, post(o))$ considerably, because now only one rule can activate a particular literal.

In order to simulate the parallel behavior of conditional effects, we have to break them up into individual operators that are executed sequentially. This means that for each conditional effect of an operator we introduce two new operators. One simulates the successful application of the rule, the other one simulates the "blocking" situation of the rule. At least one of these operators must be executed for each conditional effect in the original operator. This is something we can force by additional literals that are added to control the execution of operators. All in all this leads to a sequence of operators that has length bounded by the number of conditional effects in the original operator.

If we want to simulate the parallel behavior by a sequence of unconditional operators, the effects of the unconditional operators should not directly influence the state description, but the effect should be deferred until all operators corresponding to the set of conditional effects have been executed. For this reason, we will use a sequence of "copying operators" which copy the activated effects to the state description after all "conditional operators" have been executed. These "copying operators" can also be used to check that the set of activated effects is consistent.

**Theorem 20** *$\mathcal{S}_{\mathcal{BIC}}$ can be compiled to $\mathcal{S}_{\mathcal{BI}}$ in polynomial time preserving plan size polynomially.*

**Proof.** Assume that $\Xi = \langle \Sigma, \mathbf{O} \rangle$ is the $\mathcal{S}_{\mathcal{BIC}}$ source domain structure and assume further, without loss of generality (using a polynomial transformation), that all operators have the form

$$o_i = \langle \mathit{pre}(o_i), \{\varphi_{i,1} \Rightarrow l_{i,1}, \ldots, \varphi_{i,m_i} \Rightarrow l_{i,m_i}\}\rangle,$$

with $\varphi_{i,j} \in \mathcal{L}_\Sigma$, $l_{i,j} \in \widehat{\Sigma}$, and $l_{i,j} \neq l_{i,k}$ for $i \neq k$.

Let $\Sigma_+$ and $\Sigma_-$ be disjoint copies of $\Sigma$, which are used to record the active effects of conditional effects, and let $\Sigma_\#$ be another disjoint copy, which is used to record that an active effect has not been copied yet. Further, let $\Sigma_\mathbf{O} = \{p_o \mid o \in \mathbf{O}\}$ be a new set of atoms corresponding one-to-one to the operators in $\mathbf{O}$ and let $\Upsilon$ be a set of symbols corresponding one-to-one to all conditional effects in $\mathbf{O}$, i.e.,

$$\Upsilon = \{x_{i,j} | (\varphi_{i,j} \Rightarrow l_{i,j}) \in \mathit{post}(o_i), o_i \in \mathbf{O}\}.$$

Finally, let $c$ be a fresh atom not appearing in $\Sigma \cup \Sigma_+ \cup \Sigma_- \cup \Sigma_\# \cup \Sigma_\mathbf{O}$ that signals that copying the active effects to the state specification is in progress. The set of symbols $\Sigma'$ for the compiled domain structure is then

$$\Sigma' = \Sigma \cup \Sigma_+ \cup \Sigma_- \cup \Sigma_\# \cup \Sigma_\mathbf{O} \cup \Upsilon \cup \{c\}.$$

For each operator $o_i \in \mathbf{O}$, the compilation scheme introduces a number of new operators. The first operator we introduce is one which checks whether the conditional effects of the previous operators have all been executed, no copying is in progress and the precondition is satisfied. If this is the case, the execution of the conditional effects for this operator is started:

$$o_i^{\mathit{pre}} = \langle \mathit{pre}(o_i) \cup \neg\Sigma_\mathbf{O} \cup \{\neg c\},\ \{p_{o_i}\} \cup \neg\Sigma_+ \cup \neg\Sigma_- \cup \neg\Upsilon \cup \neg\Sigma_\#\rangle.$$

This operator enables all the "conditional effect operators." For the activated effects, we introduce the following operators:

$$o_i^{\Rightarrow,i,j} = \langle \{p_{o_i} \wedge \varphi_{i,j}\},\ \{x_{i,j}\} \cup \{p_+, p_\# \mid p = l_{i,j}\} \cup \{p_-, p_\# \mid \neg p = l_{i,j}\}\rangle.$$

In words, if the effect condition is entailed, then the activated positive or negative effect as well as the fact that the rule has been tried is recorded.

Since there is at most one effect literal for each conditional effect, a conditional effect is "blocked" if the negation of the effect condition is entailed by the state specification. For all "blocked conditional effects" we introduce the following operators:

$$o_i^{\neg,i,j} = \langle \{p_{o_i} \wedge \neg\varphi_{i,j}\},\ \{x_{i,j}\}\rangle.$$

In order to check that all conditional effects have been tried (activating the corresponding effect or not activating it because the conditional effect is blocked), the following operator is used:

$$o_i^e = \langle \{p_{o_i}\} \cup \{x_{i,j} \in \Upsilon \mid (\varphi_{i,j} \Rightarrow l_{i,j}) \in \mathit{post}(o_i)\},\ \{c\} \cup \{\neg p_{o_i}\}\rangle.$$

This operator enables copying of the activated effects to the state specification, which is achieved with the following set of operators for each atom $p \in \Sigma$:

$$\begin{array}{rcl} o_+^p & = & \langle \{c, p_+, \neg p_-, p_\#\}, \{p, \neg p_\#\}\rangle, \\ o_-^p & = & \langle \{c, \neg p_+, p_-, p_\#\}, \{\neg p, \neg p_\#\}\rangle, \\ o_\perp^p & = & \langle \{c, p_+, p_-, p_\#\}, \perp\rangle. \end{array}$$

Finally, we need an operator that checks that all possible effects have been copied. This operator also starts the "execution cycle" again by enabling the execution of another "precondition operator:"

$$o^c = \langle \{c\} \cup \neg\Sigma_\#,\ \{\neg c\}\rangle.$$

Using these definitions, we can now specify the set of compiled operators:

$$\begin{array}{rl} \mathbf{O}' = & \{o_i^{\mathit{pre}}, o_i^e \mid o_i \in \mathbf{O}\} \cup \\ & \{o^{\Rightarrow,i,j} \mid o_i \in \mathbf{O}, (\varphi_{i,j} \Rightarrow l_{i,j}) \in \mathit{post}(o_i)\} \cup \\ & \{o^{\neg,i,j} \mid o_i \in \mathbf{O}, (\varphi_{i,j} \Rightarrow l_{i,j}) \in \mathit{post}(o_i)\} \cup \\ & \{o_+^p, o_-^p, o_\perp^p \mid p \in \Sigma\} \cup \\ & \{o^c\}. \end{array}$$

Based on that, we specify a compilation scheme $\mathbf{f} = \langle f_\xi, f_i, f_g, t_i, t_g\rangle$ as follows:

$$\begin{array}{rcl} f_\xi(\Xi) & = & \langle \Sigma', \mathbf{O}'\rangle \\ f_i(\Xi) & = & \neg\Sigma_+ \cup \neg\Sigma_- \cup \neg\Sigma_\# \cup \neg\Sigma_{\mathbf{O}} \cup \neg\Upsilon \cup \{\neg c\}, \\ f_g(\Xi) & = & \neg\Sigma_{\mathbf{O}} \cup \{\neg c\}, \\ t_i(\Sigma, \mathbf{I}) & = & \mathbf{I}, \\ t_g(\Sigma, \mathbf{G}) & = & \mathbf{G}. \end{array}$$

The scheme $\mathbf{f}$ obviously satisfies conditions (2) and (3) for compilation schemes and all the functions can be computed in polynomial time. Further, $F(\Pi)$ is a $\mathcal{S}_{\mathcal{BI}}$-instance if $\Pi$ is a $\mathcal{S}_{\mathcal{BIC}}$-instance.

Let now $S \in \widehat{\Sigma}$ be a legal $\mathcal{S}_{\mathcal{BIC}}$ state specification and let $S' = R(S, o_i)$ for some operator $o_i \in \mathbf{O}$. From the above discussion, it is clear that if $S' \not\models \perp$, then there exists a sequence $\Delta'$ of operators from $\mathbf{O}'$ consisting of $o_i^{\mathit{pre}}$, followed by operators of the form $o_i^{\Rightarrow,i,j}$ and $o_i^{\neg,i,j}$ followed by the operator $o_i^e$, followed in turn by operators $o_x^p$, followed finally by the operator $o^c$, such that

$$S' = \mathit{Res}(S \cup f_i(\Xi), \Delta') \cap \widehat{\Sigma}.$$

Conversely, if $S' \models \bot$, then there does not exist any plan that transforms

$$R(S \cup f_i(\Xi), o_i^{pre})$$

into a legal state specification that contains $\neg c$ and $\neg p_{o_i}$.

Using induction on the plan length, it follows from the arguments above that there exists a plan $\Delta$ for $\Pi$ iff there exists a plan $\Delta'$ for $F(\Pi)$ and for every such plan we have $||\Delta'|| \leq ||\Delta|| \times (3 + 2m)$, with $m$ being the maximum number of conditional effects in operators of $\mathbf{O}$. Hence $\mathbf{f}$ is a polynomial-time compilation scheme preserving plan size polynomially. ■

An immediate consequence of this theorem is that $\mathcal{S}_{\mathcal{BIC}}$ and $\mathcal{S}_{\mathcal{BI}}$ form an equivalence class with respect to compilability preserving plan size polynomially.

**Corollary 21** *$\mathcal{S}_{\mathcal{BIC}}$ and $\mathcal{S}_{\mathcal{BI}}$ are polynomial-time compilable to each other preserving plan size polynomially.*

Further, we know from Corollary 15 that this class cannot become larger.

As in the case of compiling $\mathcal{S}_{\mathcal{LIC}}$ to $\mathcal{S}_{\mathcal{LC}}$, however, the result depends on the semantics chosen for executing conditional effects over partial state specifications. If we use the alternative semantics where the resulting state specification is legal when the application of all state-transformation functions leads to a theory that can be represented as a set of literals, it seems likely that there exists another scheme that preserves plan size polynomially. However, if we use the alternative semantics that deletes all the literals in $\neg(P(S, \mathit{post}(o)) - A(S, \mathit{post}(o))$ if $P(S, \mathit{post}(o)$ is consistent, then it appears to be very unlikely that we are able to identify a compilation scheme that preserves plan size polynomially.

### 6.2 Compiling Conditional Effects Away for Complete State Specifications

The next compilation scheme compiles $\mathcal{S}_{\mathcal{BC}}$ to $\mathcal{S}_{\mathcal{B}}$ and $\mathcal{S}_{\mathcal{LC}}$ to $\mathcal{S}_{\mathcal{L}}$. Since we deal with complete state specification, we do not have to take care of the condition $A(S, \mathit{post}(o)) = P(S, \mathit{post}(o))$, which is always true for complete states. This makes the compilation scheme somewhat simpler. Since $\mathcal{S}_{\mathcal{L}}$ does not allow for general boolean formulae, the scheme becomes a little bit more difficult. In general, however, the compilation scheme we will specify is very similar to the one given in the proof of Theorem 20.

**Theorem 22** *$\mathcal{S}_{\mathcal{BC}}$ can be compiled to $\mathcal{S}_{\mathcal{B}}$ and $\mathcal{S}_{\mathcal{LC}}$ can be compiled to $\mathcal{S}_{\mathcal{L}}$ in polynomial time preserving plan size polynomially.*

**Proof.** As in the proof of Theorem 20, we assume that $\Xi = \langle \Sigma, \mathbf{O} \rangle$ is the ($\mathcal{S}_{\mathcal{BC}}$ or $\mathcal{S}_{\mathcal{LC}}$) source domain structure. Further, we assume that all operators have the form

$$o_i = \langle \mathit{pre}(o_i), \{\Gamma_{i,1} \Rightarrow l_{i,1}, \ldots, \Gamma_{i,m_i} \Rightarrow l_{i,m_i}\}\rangle,$$

with $l_{i,j} \in \widehat{\Sigma}$ and $\Gamma_{i,j} \subseteq \mathcal{L}_\Sigma$ if $\Xi$ is a $\mathcal{S}_{\mathcal{BC}}$ structure or $\Gamma_{i,j} \subseteq \widehat{\Sigma}$ if $\Xi$ is a $\mathcal{S}_{\mathcal{LC}}$ structure. This means that we do not assume the effects to be unique for each conditional effect.

In addition, we assume the same set symbols for the compiled domain structure as in the proof of Theorem 20:

$$\Sigma' = \Sigma \cup \Sigma_+ \cup \Sigma_- \cup \Sigma_\# \cup \Sigma_{\mathbf{O}} \cup \Upsilon \cup \{c\}.$$

For each operator $o_i \in \mathbf{O}$, we introduce the operators $o_i^{pre}$, $o_i^e$, $o_+^p$, $o_-^p$, $o_\perp^p$, and $o^c$ as in the proof of Theorem 20. In addition, the following operators are needed:

$$\begin{aligned} o_i^{\Rightarrow,i,j} &= \langle \{p_{o_i}\} \cup \Gamma_{i,j},\ \{x_{i,j}\} \cup \{p_+, p_\# \mid p = l_{i,j}\} \cup \{p_-, p_\# \mid \neg p = l_{i,j}\}\rangle, \\ o_i^{\neg,i,j,m} &= \langle \{p_{o_i}\} \cup \{\neg\varphi_{i,j,m} \mid \varphi_{i,j,m} \in \Gamma_{i,j}\},\ \{x_{i,j}\}\rangle. \end{aligned}$$

The compiled set of operators $\mathbf{O}'$ contains all of the above operators and the compilation scheme is identical to the scheme presented in the proof of Theorem 20. This means that the only significant difference to the compilation scheme presented in the proof of Theorem 20 is the operator scheme $o_i^{\neg,i,j,m}$ which tests for each rule whether it contains an effect condition that blocks the rule. Since we have complete state specifications, every conditional effect is either activated or blocked, and the $x_{i,j}$'s are used to record that the execution of each conditional effect has been tried.

Using now similar arguments as in the proof of Theorem 20, it follows that this compilation scheme is indeed a scheme that leads to the claim made in the theorem. ■

It follows that $\mathcal{S}_{\mathcal{BC}}$ and $\mathcal{S}_{\mathcal{B}}$ are equivalent with respect to $\preceq_p^p$ and all formalisms in $[\mathcal{S}_{\mathcal{LIC}}]$ and $[\mathcal{S}_{\mathcal{LI}}]$ are equivalent with respect to $\preceq_p^p$. These two sets could be merged into one equivalence class, provided we are able to prove that, e.g., $\mathcal{S}_{\mathcal{B}}$ can be compiled to $\mathcal{S}_{\mathcal{L}}$.

### 6.3 Compiling Boolean Formulae Away

In Section 5.5 we showed that it is impossible to compile boolean formulae to conditional effects if plans are only allowed to grow linearly. However, we also sketched already the idea of a compilation scheme that preserves plan size polynomially. Here we will now show that we can compile boolean formulae to $\mathcal{S}_{\mathcal{L}}$, which is expressively equivalent to basic STRIPS, i.e., we can compile boolean formulae away completely.

**Theorem 23** *$\mathcal{S}_{\mathcal{B}}$ is polynomial-time compilable to $\mathcal{S}_{\mathcal{L}}$ preserving plan size polynomially.*

**Proof.** Assume that $\Xi = \langle \Sigma, \mathbf{O}\rangle$ is a $\mathcal{S}_{\mathcal{B}}$ domain structure. Further assume without loss of generality that all operators $o_i \in \mathbf{O}$ are of the form $o_i = \langle \varphi_i, L_i\rangle$, with $L_i \subseteq \widehat{\Sigma}$ and $\varphi_i \in \mathcal{L}_\Sigma$ (i.e., we have just one formula as the precondition instead of a set of formulae).

Let $\Sigma_\Psi$ and $\Sigma'_\Psi$ be two new sets of atoms corresponding one-to-one to all sub-formulae occurring in preconditions of operators in $\mathbf{O}$. These new atoms are denoted by $q_\psi$ and $q'_\psi$ for the sub-formula $\psi$. Atoms of the form $q'_\psi$ are used to record that the truth-value of the sub-formula $\psi$ has been computed and the atoms of the form $q_\psi$ are used to store the computed truth-value.

For each operator $o_i = \langle \varphi_i, L_i\rangle$, we will have in the target operator set the following operator:

$$o'_i = \langle \{q_{\varphi_i}, q'_{\varphi_i}\}, L_i \cup \neg\Sigma'_\Psi\rangle.$$

The set of all operators generated in this way is denoted by $\mathbf{O}'$.

Further, for each atom $p \in \Sigma$, we introduce the following two operators:

$$\begin{aligned} o_p^+ &= \langle \{p\},\ \{q'_p, q_p\}\rangle, \\ o_p^- &= \langle \{\neg p\},\ \{q'_p, \neg q_p\}\rangle. \end{aligned}$$

The set of operators generated in this way is denoted by $\mathbf{O}_\Sigma$.

For each sub-formula occurring in preconditions of $\mathbf{O}$ of the form $\psi = \psi_1 \wedge \psi_2$ the following operators are introduced:

$$\begin{aligned} o_\psi^+ &= \langle \{q'_{\psi_1}, q'_{\psi_2}, q_{\psi_1}, q_{\psi_2}\}, \{q'_\psi, q_\psi\}\rangle, \\ o_\psi^{-1} &= \langle \{q'_{\psi_1}, \neg q_{\psi_1}\}, \{q'_\psi, \neg q_\psi\}\rangle, \\ o_\psi^{-2} &= \langle \{q'_{\psi_2}, \neg q_{\psi_2}\}, \{q'_\psi, \neg q_\psi\}\rangle. \end{aligned}$$

For sub-formulae $\psi = \psi_1 \vee \psi_2$, the following operators are introduced:

$$\begin{aligned} o_\psi^{+1} &= \langle \{q'_{\psi_1}, q_{\psi_1}\}, \{q'_\psi, q_\psi\}\rangle, \\ o_\psi^{+2} &= \langle \{q'_{\psi_2}, q_{\psi_2}\}, \{q'_\psi, q_\psi\}\rangle, \\ o_\psi^{-} &= \langle \{q'_{\psi_1}, q'_{\psi_2}, \neg q_{\psi_1}, \neg q_{\psi_2}\}, \{q'_\psi, \neg q_\psi\}\rangle. \end{aligned}$$

Finally, for $\psi = \neg\gamma$, we have the following operators:

$$\begin{aligned} o_\psi^+ &= \langle \{q'_\gamma, \neg q_\gamma\}, \{q'_\psi, q_\psi\}\rangle, \\ o_\psi^- &= \langle \{q'_\gamma, q_\gamma\}, \{q'_\psi, \neg q_\psi\}\rangle. \end{aligned}$$

The set of operators generated by sub-formulae is denoted by $\mathbf{O}_\Psi$.

Now we can specify the compilation scheme $\mathbf{f}$:

$$\begin{aligned} f_\xi(\Xi) &= \langle \Sigma \cup \Sigma_\Psi \cup \Sigma'_\Psi,\ \mathbf{O}' \cup \mathbf{O}_\Sigma \cup \mathbf{O}_\Psi\rangle, \\ f_i(\Xi) &= \neg\Sigma'_\Psi, \\ f_g(\Xi) &= \neg\Sigma'_\Psi, \\ t_i(\Sigma, \mathbf{I}) &= \mathbf{I}, \\ t_g(\Sigma, \mathbf{G}) &= \mathbf{G}. \end{aligned}$$

From the construction it is obvious that all the functions are polynomial-time computable, that the state-translation functions are modular, that the induced function $F$ is a reduction, and that for every plan $\Delta$ for a source planning instance $\Pi$ there exists a plan $\Delta'$ for $F(\Pi)$ such that $||\Delta'|| \leq ||\Delta|| \times (m+1)$, with $m$ being the maximum number of sub-formulae of preconditions in $\mathbf{O}$. From that, the claim follows. ■

There might be the question whether compiling boolean formulae away could be done more efficiently. Using the result that boolean expressions can be evaluated by circuits with logarithmic depth, this should be indeed possible. However, we are satisfied here with the result that there is a compilation scheme preserving plan size polynomially at all. This result together with Theorem 22 settles the question for compilation schemes preserving plan size polynomially for all pairs of formalisms.

**Corollary 24** *All formalisms $\mathcal{X}$ with $\mathcal{X} \sqsubseteq \mathcal{S}_{\mathcal{LIC}}$ or $\mathcal{X} \sqsubseteq \mathcal{S}_{\mathcal{BC}}$ are polynomial-time compilable to each other preserving plan size polynomially.*

### 6.4 Parallel Execution Models and the Feasibility of Compilation Schemes Preserving Plan Size Polynomially

While compilation schemes that preserve plan size exactly or linearly seem to be of immediate use, a polynomial growth of the plan appears to be of little practical interest. Considering the practical experience that planning algorithms can roughly be characterized by their property of how many steps they can plan without getting caught by the combinatorial explosion and the fact that this number is significantly smaller than 100, polynomial growth does not seem to make much sense.

If we take GRAPHPLAN (Blum & Furst, 1997) into consideration again—the planning system that motivated our investigation in the first place—it turns out that this system allows for the *parallel execution* of actions. Although parallel execution might seem to add to the power of the planning system considerably, it does not affect our results at all. If a sequential plan can solve a planning instance with $n$ steps, a parallel plan will also need at least $n$ actions. Nevertheless, although the size of a plan (measured in the number of operations) might be the same, the number of time steps may be considerably smaller—which might allow for a more efficient generation of the plan. Having a look at the compilation scheme that compiles conditional effects away, it seems to be the case that a large number of generated actions could be executed in parallel—in particular those actions that simulate the conditional effects.

However, the semantics of parallel execution in GRAPHPLAN is quite restrictive. If one action adds or deletes an atom that a second action adds or deletes or if one action deletes an atom that a second action has in its precondition, then these two actions cannot be executed in parallel in GRAPHPLAN. With this restriction, it seems to be impossible to compile conditional effects away preserving the number of time steps in a plan. However, a compilation scheme that preserves the number of time steps linearly seems to be possible. Instead of such a compilation scheme, the approaches so far either used an exponential translation (Gazen & Knoblock, 1997) or modified the GRAPHPLAN-algorithm in order to handle conditional effects (Anderson et al., 1998; Koehler et al., 1997; Kambhampati et al., 1997). These modifications involve changes in the semantics of parallel execution as well as changes in the search procedure. While all these implementations are compared with the straightforward translation Gazen and Knoblock (1997) used, it would also be interesting to compare them with a compilation scheme based on the ideas spelled out in Theorem 22 as the base line.

## 7. Summary and Discussion

Motivated by the recent approaches to extend the GRAPHPLAN algorithm (Blum & Furst, 1997) to deal with *more expressive planning formalisms* (Anderson et al., 1998; Gazen & Knoblock, 1997; Kambhampati et al., 1997; Koehler et al., 1997), we asked what the term *expressive power* could mean in this context. One reasonable intuition seems to be that the term *expressive power* refers to how concisely domain structures and the corresponding plans can be expressed. Based on this intuition and inspired by recent approaches in the area of knowledge compilation (Gogic et al., 1995; Cadoli et al., 1996; Cadoli & Donini, 1997), we introduced the notion of *compilability* in order to measure the relative expressiveness of planning formalisms. The basic idea is that a *compilation scheme* can only transform the domain structure, i.e., the symbol set and the operators, while the initial state and the goal specification are not transformed—modulo some small changes necessary for technical reasons. Further, we distinguish compilation schemes according to whether the plan in the target formalism has the same size (up to an additive constant), a size bounded linearly by the size of the plan in the source formalism, or a size bounded polynomially by the original planning instance and the original plan.

Although the compilability framework appears to be a straightforward and intuitive tool for measuring the expressiveness of planning formalisms, it is possible to come up with alternative measures. Bäckström (1995), for instance, proposed to use *ESP-reductions*, which are polynomial many-one reductions on planning problems that preserve the plan size exactly. However, requiring that the transformation should be polynomial-time computable seems to be overly restrictive. In particular, if we want to prove that one formalism is *not* as expressive as another one, we had better proven that there exists no compilation scheme regardless of how much computational resources the compilation process may need. Furthermore, there appear to be severe technical problems to using Bäckström's (1995) framework for proving negative results. On the other hand, all of the positive results reported by Bäckström are achievable in the compilation framework because the transformations he used are in fact compilation schemes. Taking all this together, it appears to be the case that the compilation framework is superior from an intuitive and technical point of view.

Another approach to judging the expressiveness of planning formalisms has been proposed by Erol and colleagues (1994, 1996). They measure the expressiveness of planning formalisms according to the set of plans a planning instance can have. While this approach contrasts *hierarchical task network* planning nicely with STRIPS-planning, it does not help us in making distinctions between the formalisms in the $\mathcal{S}$-family.

The compilability framework is mainly a theoretical tool to measure how concisely domain structures and plans can be expressed. However, it also appears to be a good measure of how difficult planning becomes when a new language feature is added. *Polynomial-time* compilation schemes that preserve the plan size linearly indicate that it is easy to integrate the feature that is compiled away. One can either use the compilation scheme as is or mimic the compilation scheme by extending the planning algorithm. If only a polynomial-time compilation scheme leading to a polynomial growth of the plan is possible, then this is an indication that adding the new feature requires most probably a significant extension of the planning algorithm. If even a compilation scheme preserving plan size polynomially can be ruled out, then there is most probably a serious problem integrating the new feature.

Using this framework, we analyzed a large family of planning formalisms ranging from basic STRIPS to formalisms with conditional effects, boolean formulae, and incomplete state specifications. The most surprising result of this analysis is that we are able to come up with a complete classification. For each pair of formalisms, we were either able to construct a *polynomial-time compilation scheme* with the required size bound on the resulting plans or we could prove that compilation schemes are impossible—even if the computational resources for the compilation process are unbounded. In particular, we showed for the formalisms considered in this paper:

- incomplete state specifications and literals in preconditions can be compiled to basic STRIPS preserving plan size exactly,

- incomplete state specifications and literals in preconditions and effect conditions can be compiled away preserving plan size exactly, if we have already conditional effects,

- and there are no other compilation schemes preserving plan size linearly except those implied by the specialization relationship and those described above.

If we allow for polynomial growth of the plans in the target formalism, then all formalisms not containing incomplete state specifications and boolean formulae are compilable to each other. Incomplete state specifications together with boolean formulae, however, seem to add significantly to the expressiveness of a planning formalism, since these cannot be compiled away even when allowing for polynomial growth of the plan and unbounded resources in the compilation process.

It should be noted, however, that some of these results hold only if we use the semantics for conditional effects over partial state specifications as spelled out in Section 2.1. For other semantics, we may get slightly different results concerning the compilability of conditional effects over partial states.

One question one may ask is what happens if we consider formalisms with boolean formulae that are syntactically restricted. As indicated at various places in the paper, restricted formulae, such as CNF or DNF formulae, can sometimes be easily compiled away. However, there are also cases when this is impossible. For example, it can be shown that CNF formulae cannot be compiled to basic STRIPS preserving plan size linearly (Nebel, 1999), which confirms Bäckström's (1995) conjecture that CNF-formulae in preconditions add to the expressive power of basic STRIPS.

Another question is how reasonable our restrictions on a compilation scheme are. In particular, one may want to know whether *non-modular* state-translation functions could lead to more powerful compilation schemes. First of all, requiring that the state-translation functions are modular seems to be quite weak considering the fact that a compilation scheme should only be concerned with the domain structure and that the initial state and goal specification should not be transformed at all. Secondly, considering the fact that the state-translation functions do not depend on the operator set, more complicated functions seem to be useless. From a more technical point of view, we need modularity in order to prove that conditional effects and boolean formulae cannot be compiled away preserving plan size linearly. For the conditional effects, modularity or a similar condition seems to be crucial. For the case of boolean formulae, we could weaken the condition to the point that we require only that state-translation functions are computable by circuits of constant depth—or something similar. In any case, the additional freedom one gets from non-modular state-translation functions does not seem to be of any help because these functions do not take the operators into account. Nevertheless, it seems to be an interesting theoretical problem to prove that more powerful state-translation functions do not add to the power of compilation schemes.

Although the paper is mainly theoretical, it was inspired by the recent approaches to extend the GRAPHPLAN algorithm to handle more powerful planning formalisms containing conditional effects. So, what are the answers we can give to open problems in the field of planning algorithm design? First of all, Gazen and Knoblock's (1997) approach to compiling conditional effects away is optimal if we do not want to allow plan growth more than by a constant factor. Secondly, all of the other approaches (Anderson et al., 1998; Kambhampati et al., 1997; Koehler et al., 1997) that modify the GRAPHPLAN algorithm are using a strategy similar to a polynomial-time compilation scheme preserving plan size polynomially. For this reason, these approaches should be compared to a "pure compilation approach" using the ideas from the compilation scheme developed in the proof of Theorem 22 as the base line. Thirdly, allowing for unrestricted boolean formulae adds again a level of expressivity because they cannot be compiled away with linear growth of the plan size. In fact, approaches such as the one by Anderson and colleagues (1998) simply expand the formulae to DNF accepting an exponential blow-up. Again, we cannot do better than that if plan size should be preserved linearly. Fourthly, if we want to add partial state specifications on top of general boolean formulae, this would amount to an increase of expressivity that is much larger than adding conditional effects or general formulae to basic STRIPS, because in this case there is no way to compile this away even if we allow for polynomial plan growth.

Finally, one may wonder how our results apply to planning approaches that are based on translating (bounded) planning problems to propositional logic such as SATPLAN (Kautz & Selman, 1996) or BLACKBOX (Kautz & Selman, 1998). Since the entire analysis of the relative expressiveness of planning formalisms uses the assumption that we compile from one planning formalism to another planning formalism, the results do not tell us anything about the size of representations if we switch to another formalism. In particular, it seems possible to find an encoding of (bounded) planning problems with conditional operators in propositional logic which is as concise as an encoding of unconditional operators. The only advice our results give is that such a concise encoding will not be found by first translating conditional actions to unconditional actions and then using the "standard" encoding for unconditional actions (Kautz, McAllester, & Selman, 1996) to generate boolean formulae. However, addressing the problem of determining the conciseness of representation in this context appears to be an interesting and relevant topic for future research.

## Acknowledgments

The research reported in this paper was started and partly carried out while the author enjoyed being a visitor at the AI department of the University of New South Wales. Many thanks go to Norman Foo, Maurice Pagnucco, and Abhaya Nayak and the rest of the AI department for the discussions and cappuccinos. I would also like to thank Birgitt Jenner and Jacobo Toran for some clarifications concerning circuit complexity.

## Appendix A: Symbol Index

| Symbol | Page | Explanation |
|---|---|---|
| $\|\cdot\|$ | 292 | cardinality of a set |
| $\|\|\cdot\|\|$ | 277 | size of an instance |
| $\Rightarrow$ | 274 | symbol used in conditional effects |
| $\sqsubseteq$ | 279 | syntactic specialization relation |
| $\preceq^x_y$ | 282 | compilability relation with restriction $x$ and $y$ |
| $\bot$ | 273 | boolean constant denoting falsity, also denoting the illegal state specification |
| $\top$ | 273 | boolean constant denoting truth |
| $a(\cdot)$ | 295 | advice function |
| $A(\cdot,\cdot)$ | 275, 275 | active effects of an operator in a state or state specification |
| $\mathsf{AC}^0$ | 298 | complexity class |
| $C$ | 298 | boolean circuit |
| $\mathbf{C}$ | 298 | family of boolean circuits |
| coNP | 272 | complexity class |
| coNP/poly | 295 | non-uniform coNP |
| $CWA_\Sigma(\cdot)$ | 284 | closing a set of literals w.r.t. $\Sigma$ |
| $\Delta$ | 277 | plan, i.e., sequence of operators |
| $\Delta^p_i$ | 295 | complexity class in the polynomial hierarchy |
| $I$ | 295 | instance of a problem |
| $\mathbf{I}$ | 277 | initial state description |
| $\mathbf{f}$ | 282 | compilation scheme ($= \langle f_\xi, f_i, f_g, t_i, t_g \rangle$) |
| $F(\cdot)$ | 282 | transformation induced by compilation scheme |
| $f_\xi, f_i, f_g$ | 282 | components of a compilation scheme |
| $\mathbf{G}$ | 277 | goal of a planning task |
| $\Gamma$ | 274 | set of boolean formulae |
| $\gamma, \varphi, \psi$ | 273 | boolean formulae |
| $l$ | 287 | literal |
| $L, K$ | 273 | sets of literals |
| $\mathcal{L}_\Sigma$ | 273 | all boolean formulae that use atoms from $\Sigma$ |
| $Mod(\cdot)$ | 274 | set of all models of a theory |
| $\mathsf{NC}^1$ | 298 | complexity class |
| $neg(\cdot)$ | 273 | negative literals in a set of literals |
| NP | 272 | complexity class |
| NP/poly | 295 | non-uniform NP |
| $o$ | 274 | operator ($= \langle$*pre*, *post*$\rangle$) |
| $\mathbf{O}$ | 276 | set of operators |
| $\mathbf{O}^\star$ | 277 | set of finite sequences of operators |
| $p, q, u, v, x$ | 274 | propositional atoms |
| $P(\cdot, \cdot)$ | 275 | potentially active effects of an operator for a given state specification |
| P | 272 | complexity class |
| P/poly | 295 | non-uniform P |
| PH | 295 | the polynomial hierarchy |
| $pos(\cdot)$ | 273 | positive literals in a set of literals |
| PLANEX | 279 | plan existence problem |
| *post* | 274 | postconditions of an operator |
| *pre* | 274 | preconditions of an operator |
| PSPACE | 272 | complexity class |
| $\Pi$ | 277 | planning instance ($= \langle \Xi, \mathbf{I}, \mathbf{G} \rangle$) |
| $\Pi_i^p$ | 295 | complexity class in the polynomial hierarchy |
| $R(\cdot, \cdot)$ | 276 | maps a state specification and an operator to a new state |
| $Res(\cdot, \cdot)$ | 277 | extension of $R(\cdot, \cdot)$ to plans |
| $s$ | 274 | state (or truth assignment) |
| $S$ | 274 | state specification |
| $\mathcal{S}$ | 278 | the STRIPS planning formalism |
| $\mathcal{S}_\mathcal{L}$ | 278 | STRIPS with literals in preconditions |
| $\mathcal{S}_\mathcal{B}$ | 278 | STRIPS with boolean formulae in preconditions |
| $\mathcal{S}_\mathcal{I}$ | 278 | STRIPS with incomplete state descriptions |
| $\mathcal{S}_\mathcal{C}$ | 278 | STRIPS with conditional effects |
| $\mathcal{S}_{\ldots}$ | 278 | STRIPS with combinations of the above extensions |
| $[\mathcal{S}_{\mathcal{L}\mathcal{I}\mathcal{C}}]$ | 283 | equivalence classes induced by $\preceq_p^1$ |
| $[\mathcal{S}_{\mathcal{L}\mathcal{I}}]$ | 283 | equivalence classes induced by $\preceq_p^1$ |
| $\sigma(\cdot)$ | 273 | all propositional atoms used in a set of literals |
| $\mathbf{\Sigma}$ | 273 | countably infinite set of propositional atoms |
| $\Sigma$ | 273 | finite subset of $\mathbf{\Sigma}$ |
| $\widehat{\Sigma}$ | 273 | set of literals overs $\Sigma$ |
| $\Sigma_i^p$ | 295 | complexity class in the polynomial hierarchy |
| $t_i, t_g$ | 282 | state-translation functions in a compilation scheme |
| $\mathbf{U}$ | 292 | universal literals |
| $w$ | 298 | a word over $\{0,1\}^\star$ |
| $\mathsf{X}$ | 295 | some complexity class |
| $\mathcal{X}, \mathcal{Y}$ | 272 | some planning formalisms |
| $\Xi$ | 277 | domain structure ($= \langle \Sigma, \mathbf{O} \rangle$) |
| $\mathbf{\Xi}$ | 299 | family of domain structures |